\documentclass[10pt,twocolumn,letterpaper]{article}

\usepackage{iccv}              % To produce the CAMERA-READY version

\usepackage{graphicx}
\usepackage{subcaption}
\usepackage{multirow}
\usepackage{colortbl}
\usepackage{mathtools}

\usepackage[ruled,linesnumbered]{algorithm2e}
\newtheorem{theorem}{Theorem}
\newtheorem{definition}{Definition}%
\definecolor{iccvblue}{rgb}{0.21,0.49,0.74}
\usepackage[pagebackref,breaklinks,colorlinks,allcolors=iccvblue]{hyperref}

\title{Causal Prompt Calibration Guided Segment Anything Model for Open-Vocabulary Multi-Entity Segmentation}

\author{Jingyao Wang*, Jianqi Zhang*, Wenwen Qiang, Changwen Zheng \\
University of Chinese Academy of Sciences\\
Institute of Software Chinese Academy of Sciences
}
\begin{document}
\maketitle

% \begin{document}
% \twocolumn[{%
% \renewcommand\twocolumn[1][]{#1}%
% \maketitle
% \begin{center}
%     \includegraphics[width=\textwidth]{fig/intro_new.pdf}
%     % \vspace{-0.2in}
%     \captionof{figure}{Segmentation results of unseen classes on ImageNet \cite{deng2009imagenet} and real-life deep-sea samples (OceanCOCO). All classes in OceanCOCO are merged with the classes within ImageNet to form the vocabulary, and the randomly selected segmentation results are visualized.}
%     \label{fig:vis_img}
%     \vspace{0.1in}
% \end{center}%
% }]

\begin{abstract}
% \vspace{-0.1in}
Despite the strength of the Segment Anything Model (SAM), it struggles with generalization issues in open-vocabulary multi-entity segmentation (OVMS). Through empirical and causal analyses, we find that (i) the prompt bias is the primary cause of the generalization issues; (ii) this bias is closely tied to the task-irrelevant generating factors within the prompts, which act as confounders and affect generalization. To address the generalization issues, we aim to propose a method that can calibrate prompts to eliminate confounders for accurate OVMS. Building upon the causal analysis, we propose that the optimal prompt for OVMS should contain only task-relevant causal factors. We define it as the causal prompt, serving as the goal of calibration. Next, our theoretical analysis, grounded by causal multi-distribution consistency theory, proves that this prompt can be obtained by enforcing segmentation consistency and optimality. Inspired by this, we propose CPC-SAM, a Causal Prompt Calibration method for SAM to achieve accurate OVMS. It integrates a lightweight causal prompt learner (CaPL) into SAM to obtain causal prompts. Specifically, we first generate multiple prompts using random annotations to simulate diverse distributions and then reweight them via CaPL by enforcing causal multi-distribution consistency in both task and entity levels. To ensure obtaining causal prompts, CaPL is optimized by minimizing the cumulative segmentation loss across the reweighted prompts to achieve consistency and optimality. A bi-level optimization strategy alternates between optimizing CaPL and SAM, ensuring accurate OVMS. Extensive experiments validate its superiority. The code will be available at \url{https://github.com/WangJingyao07/CPC-SAM}.

\end{abstract}    
\vspace{-0.15in}

\section{Introduction}
\label{sec:1}

Open-vocabulary multi-entity segmentation (OVMS), a crucial segmentation task, seeks to group pixels in an image into regions with labels of different semantic classes, where the labels come from an open-vocabulary set \cite{Zhang_2023_ICCV,DBLP:journals/corr/abs-2001-05566}. 
Unlike traditional segmentation that uses the same set of labels in both training and testing phases, OVMS introduces new classes in the testing phase that were unseen before \cite{10420487,sun20243d}. Meanwhile, each sample may contain multiple entities belonging to different classes, e.g., ``nose'' and ``eyes'' in a facial image (\textbf{Figure \ref{fig:ex_3_2}}), further complicating segmentation.

Recently, Segment Anything Model (SAM) \cite{sam} has become an effective tool for solving segmentation tasks. 
However, our experimental results show that SAM performs poorly on OVMS compared to general segmentation tasks.
Specifically, we construct a custom dataset called OceanCOCO (Appendix \ref{sec_app:dataset}), which extends the COCO dataset \cite{lin2014microsoft} with 46 real-world deep-sea species to simulate practical OVMS applications. These classes are divided into base and target classes in a ratio of 13:6 to assess generalization capabilities. We then record the performance of SAM across training epochs on both base and target classes. \textbf{Figure \ref{fig:motivation_ex_generalization}} shows that while base-class accuracy improves consistently, the performance of target classes peaks early and then sharply declines. This phenomenon highlights the limitations of SAM in generalizing to unseen classes in OVMS.

Based on the characteristics of OVMS, the failure of SAM may stem from its insufficient generalization ability to out-of-distribution (OOD) data. Specifically, in OVMS, the model may encounter a large number of classes that were unseen in the training phase. This results in a significant distribution shift between the classes in the testing and training phases, making it difficult for a model trained on base classes to generalize to unseen target classes. Upon revisiting SAM, its performance is predominantly determined by two components: the pre-trained network and the prompt. Existing studies \cite{zhou2023limaalignment,ye2025limoreasoning} generally suggest that the network has learned cross-domain feature representations from massive amounts of unlabeled data (covering 11 million images for SAM) during pre-training, which enables it to generalize to new classes. Therefore, it can be conservatively assumed that the pre-trained network of SAM is effective in handling OOD problems. 
In contrast, the bottleneck in the prompt is more pronounced, specifically: (i) the manually interactive prompts (e.g., manually marking points and bounding boxes) rely on expert knowledge and may struggle to accurately capture fine-grained attributes of unseen classes (e.g., the tentacle structures of marine organisms); (ii) the generative prompts are constrained by the quality of the generative model, potentially leading to a semantic mismatch between the prompt and the image.
To quantify the impact of prompt quality, we conduct controlled experiments (\textbf{Subsection \ref{sec:analysis_empirical}}). The results in \textbf{Figure \ref{fig:motivation_ex_prompt_bias}} show that SAM achieves a Dice score of 46.1\% with standard prompts (pre-trained model-generated) versus 78.5\% with refined prompts (expert-annotated) on the unseen classes. This 32.4\% performance gap identifies prompt bias as the primary bottleneck in the OOD generalization of SAM.

To eliminate the above prompt bias, a straightforward approach is to calibrate the generated prompts. For example, for manually interactive prompts, expert validation can filter accurate semantic regions; for model-generated prompts, we can refine them separately with additional data \cite{zhangblo,yuan2024open}. However, expert annotation is costly and impractical for large-scale applications, while model-based optimization heavily relies on data quality, and not all unseen classes have available annotated data, making prompt bias hard to avoid. 
\textbf{To address these issues, we aim to explore a way to automatically calibrate prompts without additional data. We focus on two key questions: (i) What a good prompt should be? (ii) How to obtain such a prompt without additional data?
We propose to analyze from a causal perspective with Structural Causal Models (SCMs) (\textbf{Subsection \ref{sec:analysis_causal}}).} From the data generation mechanism \cite{hu2022causal,deshpande2022deep}, samples are produced by labels and environmental effects, corresponding to a set of generating factors. In OVMS, these factors can be divided into (i) task-relevant causal factors which represent the semantics of different entities (e.g., shape and texture of ``cat''), and (ii) task-irrelevant ones for environmental effects which is useless for segmentation. The role of prompts can be seen as generating factors templates that guide SAM to extract causal factors, establishing a mapping between the sample and label. 
However, in OVMS, the model encounters numerous unseen classes in the testing phase. 
This OOD issue may make the prompts contain task-irrelevant factors, i.e., covering the background due to inaccurate annotation.
These factors acts as confounders and leads to prompt bias. 
This bias may lead SAM to incorrectly link task-irrelevant factors to labels, affecting performance.
Thus, \textbf{a good prompt should contain only accurate causal factors without confounders}, defined as \emph{causal prompt}.
Next, we turn to explore how to obtain causal prompts. 
Following causal multi-distribution consistency theory \cite{pearl2009causality}, achieving consistent results across different datasets for the same task ensures that the model learns only the causal factors while eliminating confounders.
Thus, by making prompts generated under different conditions to enforce consistency (e.g., driving losses to zero), the model ensures that the prompts preserve only invariant causal factors (\textbf{Theorem 1}).

Based on the above insights, we propose a Causal Prompt Calibration method for SAM to achieve accurate OVMS (CPC-SAM). It embeds a lightweight Causal Prompt Learner (CaPL) into SAM to obtain causal prompts. 
Specifically, we first generate multiple prompts for the same sample using random annotations to simulate diverse distributions. These prompts are then reweighted in CaPL to obtain causal prompts by enforcing causal multi-distribution consistency. Considering the multi-entity characteristics of the OVMS samples, we perform calibration across both tasks and entities.
To ensure the accuracy of reweighting, we optimize CaPL by minimizing the cumulative segmentation loss across different prompts, ensuring that the prompts capture only causal factors relevant to the labels. 
A bi-level optimization strategy is proposed to optimize CaPL and SAM, ensuring accurate OVMS. Extensive experiments validate its superiority.

The main contributions include:
\begin{itemize}
    \item Our empirical and causal analysis reveals that (i) SAM faces generalization risks in OVMS due to prompt bias; (ii) this bias is tied to the task-irrelevant generating factors, which act as confounders and affect prompt quality.
    \item We propose causal prompt calibration guided SAM (CPC-SAM) for OVMS. It embeds a lightweight network to calibrate prompts with causal multi-distribution consistency and perform optimization via a bi-level process. 
    \item Extensive experiments on various benchmark datasets demonstrate the advantages of CPC-SAM for OVMS.
\end{itemize}

\section{Related Work}
\label{sec:2}

\paragraph{Open-Vocabulary Semantic Segmentation (OVSS)} addresses the limitations of closed vocabulary sets in prior segmentation tasks. Early approaches \cite{bucher2019zero,xian2019semantic,xu2022groupvit} either used generative models to synthesize pixel-level features from unseen class, or aligned semantics and visual features using embeddings \cite{mikolov2013efficient,wang2023amsa}. Considering that their vocabulary is restricted, recent works \cite{xu2023side,ghiasi2022scaling,li2022language,liang2023open,zhou2022extract,jiao2023learning} leveraged pre-trained models like CLIP \cite{clip} and SAM \cite{sam} for segmentation. However, they rely on heavy prompt generators and large datasets, making them inefficient in inference and hard to generalize. Although several studies \cite{wang2024sam, yuan2024open, zhangblo,li2024clipsam} have attempted to enhance performance by introducing additional data, they still exhibit significant prompt bias when encountering unseen semantic categories during the OVMS testing phase. To address these issues, we aim to explore, from a causal perspective, what constitutes a good prompt and how to learn it. Based on the theoretical results and extensive analyses, we propose a method to obtain accurate prompts for OVMS without relying on external data.

\vspace{-0.15in}
\paragraph{Vision Language Models (VLMs)} are pre-trained to align image and text representations. Recent studies on VLMs \cite{clip,li2024clipsam,hafner2021clip,luo2022clip4clip,wang2024causal} have significantly enhanced the generalization ability. Among them, SAM has demonstrated remarkable performance on various tasks, e.g., grounding \cite{liu2023grounding}, tracking \cite{cheng2023segment}, and generation \cite{zhang2023personalize}. However, SAM faces generalization issues due to a mismatch between the prompt and data (demonstrated in \textbf{Section \ref{sec:analysis}}). Although some works \cite{wu2023medical,tancik2020fourier} consider using additional to fine-tune the prompt, it remain constrained by the data limitations of OVMS. Thus, this work aims to obtain more accurate \emph{causal prompts} to enhance performance.

\vspace{-0.15in}
\paragraph{Prompt Tuning} is a strategy for adapting VLMs to new tasks \cite{lester2021power,li2021prefix,liu2023pre}. For instance, CoOp \cite{zhou2022learning} and MaskCLIP \cite{bahng2022exploring,jia2022visual} incorporate learnable vectors and masks into prompts to adapt VLMs to various recognition tasks. Prompt tuning is also widely used in OVMS \cite{li2022language,du2022learning}. Unlike earlier methods \cite{liang2023open,dong2023maskclip,yang2023attentive}, we replace default prompts with learnable causal prompts. These prompts capture causal generating factors that are closely tied to semantic labels while removing irrelevant factors, e.g., avoiding the background ``sedimentary rock'' when segmenting deep-sea ``coral''. 
We are the first to apply causality to SAM, achieving robust OVMS, an improvement not previously reported.

\section{Problem Analysis and Motivation}
\label{sec:analysis}

\subsection{Problem Settings}
\label{sec:analysis_notation}

OVMS requires the model to segment images into multiple semantic regions by recognizing both seen and unseen classes, unlike traditional segmentation which is limited to predefined classes. Considering the difficulty in segmenting unseen classes, recent works \cite{yuan2024open,liang2023open,xu2023side} mainly leverage the powerful VLMs, e.g., SAM, with prompts for OVMS. Formally, given an image \( x_i \), the model needs to assign multiple labels \( y_i=\{y^1_i, \dots, y^m_i\} \) corresponding to distinct entities from an open vocabulary \( \mathcal{Y}  \) to $x_i$ with prompt $p_i$. 
This is commonly achieved by minimizing the loss $\mathcal{L}(f_\phi; \mathcal{D}) = \sum_{(x_i,p_i) \in \mathcal{D}} y_{i} \log f_\phi(x_{i}, p_i)$, where \( f_\phi \) is the segmentation model, $\mathcal{D}$ is the dataset, and \( (x_i, p_i, y_i) \) denotes the image, prompt, and label, respectively.

\begin{figure*}
    \begin{minipage}[t]{0.74\textwidth}
    % \vspace{-0.15in}
        \centering    
        \begin{subfigure}{0.32\linewidth}
        \centering
        \includegraphics[width=\textwidth]{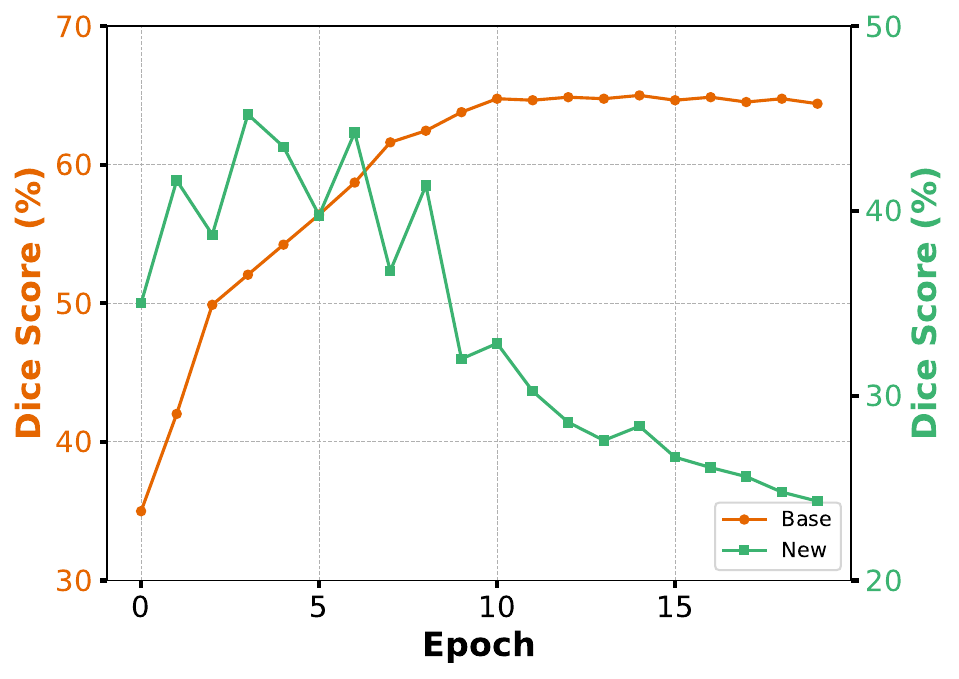}
        \caption{Existence of generalization issue} 
        \label{fig:motivation_ex_generalization}
    \end{subfigure}
    \hfill
    \begin{subfigure}{0.32\linewidth}
        \centering
        \includegraphics[width=\textwidth]{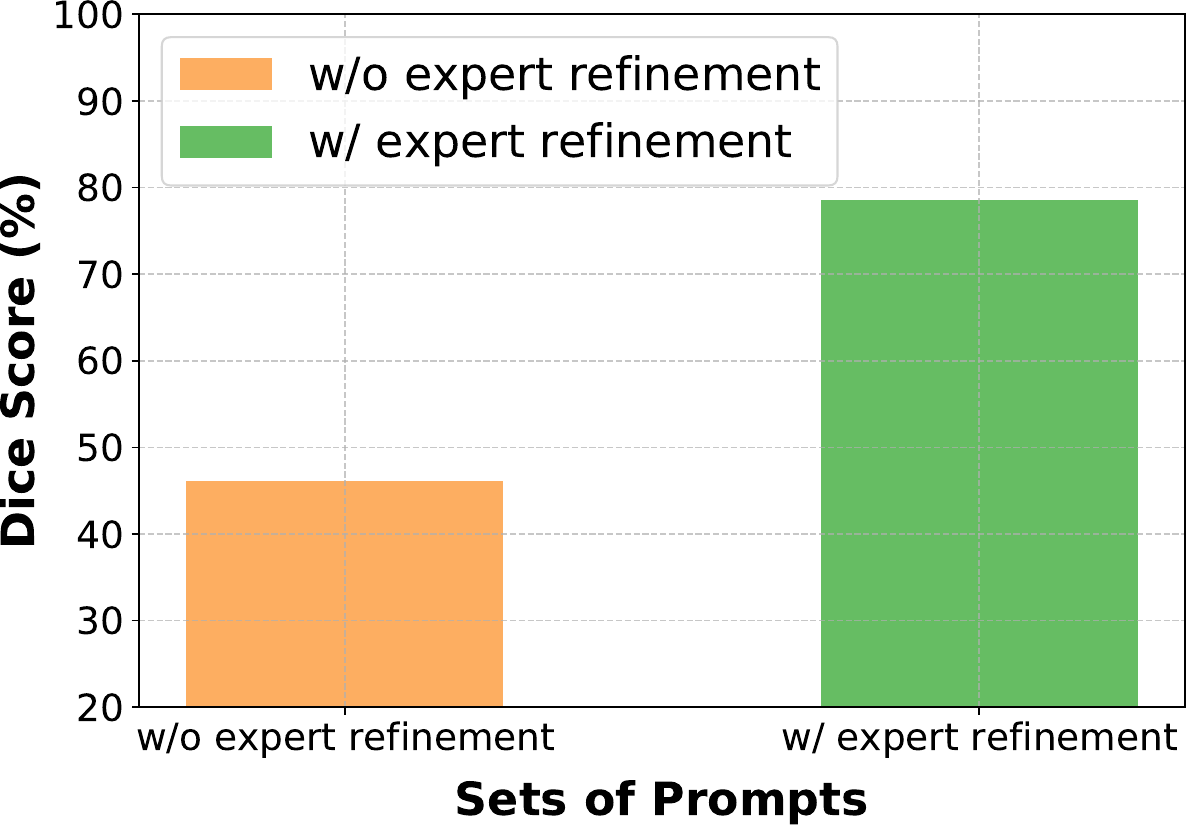}
        \caption{Existence of prompt bias} 
        \label{fig:motivation_ex_prompt_bias}
    \end{subfigure}
    \hfill
    \begin{subfigure}{0.32\linewidth}
        \centering
        \includegraphics[width=\textwidth]{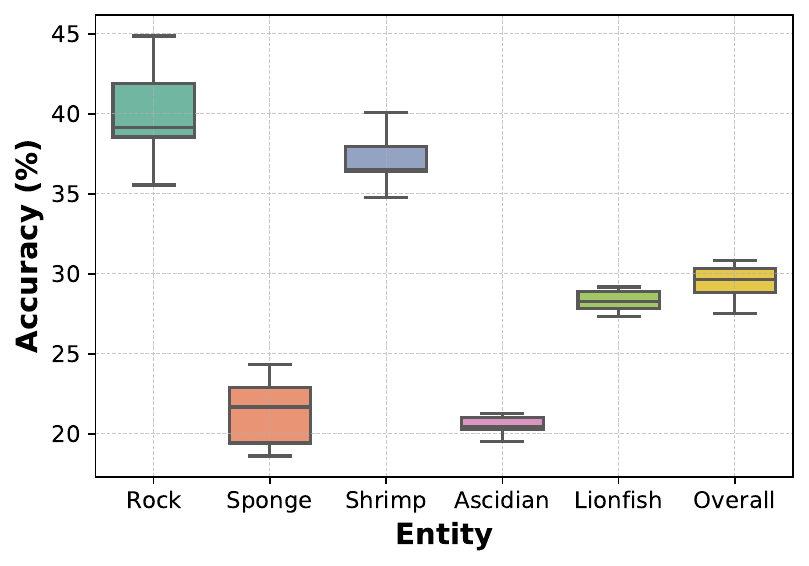}
        \caption{Prompt bias across entities}
        \label{fig:prompt_bias_entity}
    \end{subfigure}
    \vspace{-0.1in}
    \caption{Motivating experiments. \textbf{(b) Existence of generalization issue}, showing the dice score trends of training and unseen classes during training. \textbf{(c) Existence of prompt bias}, illustrating the dice score of SAM with prompts generated under different conditions. \textbf{(d) Results on prompt bias across multiple entities}, showing the impact of the same prompts on segmenting various entities.}
    \label{fig:motivating_ex}
    \vspace{-0.1in}
    \end{minipage}
    \hfill
    \begin{minipage}[t]{0.24\textwidth}
    % \vspace{-0.2in}
        \centering
        \includegraphics[width=0.88\textwidth]{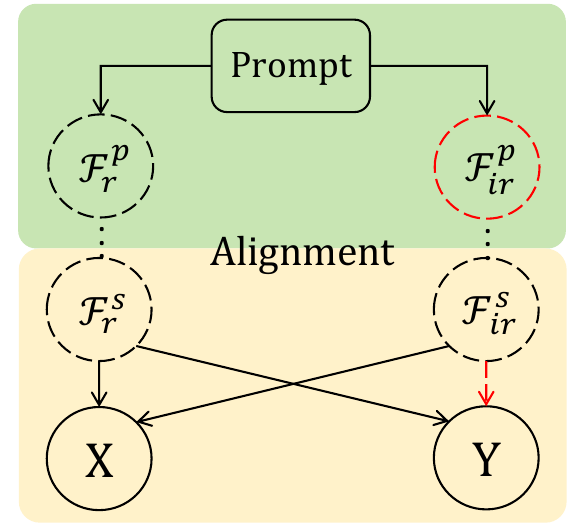}
        \vspace{-0.1in}
        \caption{SCM of the prompt-tuning process. Solid and dashed circles are observable and unobservable variables, respectively.}
        \label{fig:scm}
    \vspace{-0.1in}
    \end{minipage}
\end{figure*}

\subsection{Empirical Evidence}
\label{sec:analysis_empirical}

\paragraph{Existence of Generalization Issue} 

Based on the above settings, in OVMS, the model may encounter a large number of semantic classes that were not seen in the training phase. 
This creates a substantial distribution shift between classes in the testing and training phases, making it challenging for a model trained on base classes to generalize to unseen target classes.
As one powerful segmentation VLM, SAM \cite{sam} is considered to be adaptable well to various segmentation tasks. 
To explore whether SAM can still perform well in OVMS, we conduct a set of experiments. We select the self-collected dataset, OceanCOCO (\textbf{Appendix \ref{sec_app:dataset}}) for evaluation. OceanCOCO introduces real-world deep-sea samples, which contains rare species classes to simulate practical applications. The ratio of base classes to new classes is 13:6. We then perform fine-tuning and record model performance on both base and new classes during training, i.e., dice score (with detailed calculation in \textbf{Appendix \ref{sec_app:metric}}). As shown in \textbf{Figure \ref{fig:motivation_ex_generalization}}, performance on base classes progressively improved with increasing training epochs, while performance on new classes initially increased and then declined, illustrating generalization issue.

\vspace{-0.15in}
\paragraph{Exploration of Generalization Issue}
Given the characteristics of OVMS, the failure of SAM may be attributed to its limited ability to generalize to OOD data.
Upon revisiting SAM, its performance largely depends on the pre-trained network and prompts. The network, trained on 11 million images, is considered to learn cross-domain features that enable generalization to unseen classes in OOD tasks \cite{zhou2023limaalignment,ye2025limoreasoning}. By comparison, the aforementioned OOD generalization issue is more likely to lie in the prompts, specifically: (i) the manually interactive prompts depend on expert knowledge and may struggle to precisely capture the attributes of unseen classes; (ii) the generative prompts are constrained by the generative models, may result in a semantic mismatch between the prompt and the image.
To verify this hypothesis, we construct a controlled experiment.
We first select a test set from the unseen classes in OceanCOCO, including 60 samples covering 12 classes. Each sample may contain multiple entities.
Next, we generate two sets of prompts:
(i) The first set is generated using a fixed pre-trained prompt generator \cite{yuan2024open}, which is well-trained on the base classes of the OceanCOCO dataset.
(ii) The second set builds on the first but incorporates expert knowledge, i.e., inviting researchers to refine segmentation boundaries for each entity within the prompts.
Using these two sets of prompts, we then perform segmentation with SAM and record the results. The results in \textbf{Figure \ref{fig:motivation_ex_prompt_bias}} show that segmentation based on the second set of prompts outperforms the first by 32.4\%. This demonstrates the importance of prompt quality in OVMS. In other words, prompt quality may be the primary cause affecting the OOD generalization ability of SAM.
Additionally, considering the multi-entity nature of OVMS, we visualize segmentation results for different entities within the same sample based on the first set of prompts. \textbf{Figure \ref{fig:prompt_bias_entity}} shows that the impact of the same prompt on different entities varies. This emphasizes that we need to consider the impact of prompt bias on different entities to improve overall performance. These results demonstrate the impact of prompt bias and also suggest that a good prompt helps reduce the generalization gap.

\subsection{Causal Analysis and Motivation}
\label{sec:analysis_causal}

Building upon the above analyses, a direct way to reduce prompt bias is to calibrate the prompts. For example, expert validation can refine semantic regions for manually interactive prompts, while generative prompts can be adjusted with additional data. However, expert annotation is costly and impractical for large-scale use, and model-based optimization depends on data quality, making the bias hard to eliminate. 
To address these issues, we aim to explore a way to automatically calibrate prompts without additional data. Achieving this goal requires answering two key questions: (i) What defines a good prompt? (ii) How can such a prompt be obtained without additional data? In this section, we propose to answer these questions from a causal perspective.

\vspace{-0.2in}
\paragraph{What Defines A Good Prompt: A Causal Perspective}

To explore what a good prompt is, we construct an SCM to analyze how prompts impact the segmentation of SAM and then define what a good prompt should be. The SCM is shown in \textbf{Figure \ref{fig:scm}}. It consists of two main components: the segmentation process of SAM (yellow box) and the guidance provided by prompts (green box). 
In the segmentation process, $X$ and $Y$ represent input samples and their corresponding labels within the task. Following the data generation mechanism \cite{hu2022causal,deshpande2022deep,wang2023hacking}, the samples $X$ are considered to be produced by both (i) generating factors related to labels, i.e., the task-relevant factors $F_{r}^s$, and (ii) unobserved factors (e.g., environmental effects), i.e., the task-irrelevant factors $F_{ir}^s$. The task-relevant factors $F_{r}^s$ determine features that are highly relevant to the labels of different entities ($F_r^s\to Y $), e.g., coat colors and ear types for segmenting ``cat''. Conversely, task-irrelevant factors influence semantics that do not provide helpful information, e.g., background. Then, we get $F_r^s\to X $, $F_{ir}^s\to X $, and $F_r^s\to Y$. 
Thus, to access accurate segmentation, the model needs to capture only causal factors $F_{r}^s$ from the sample and then use them to establish the causal relationship between $X$ and $Y$. 
To guide the model in learning these causal relationships, prompts are introduced. A prompt can be regarded as a template containing causal factors $F_r^p$ associated with different entity labels, e.g., textual label and mask of ``cat''. By aligning the factors extracted from the samples with those encoded in the prompt, the model can achieve more accurate segmentation. However, in OVMS, the model may encounter numerous classes that were unseen in the training phase. \cite{Zhang_2023_ICCV,DBLP:journals/corr/abs-2001-05566,li2024segearth}. In some cases, certain classes may be extremely rare or even unseen for humans (e.g., deep-sea exploration or military scenarios), increasing the risk of mislabeling. These data limitations may introduce irrelevant factors $F_{ir}^p$ into the prompt, acting as confounders and leading to prompt bias. This bias may cause the segmentation model to mistakenly associate irrelevant factors $F_{ir}^s$ with the labels ($F_{ir}^s \to Y$), affecting generalization. To mitigate prompt bias, we propose the optimal prompt as \emph{causal prompt}, which contains only accurate causal generating factors essential for distinguishing different entity labels.

\vspace{-0.15in}
\paragraph{How To Obtain Such Prompt}

Based on the above analysis, we propose that obtaining causal prompts is essentially about selecting the causal factors in the prompt.
According to the causal multi-distribution consistency theory \cite{pearl2009causality,Cheng2017causal}, causal factors for a task remain invariant across datasets, regardless of environmental variations. This means that valid results are likely to come from these invariant causal factors while discarding environmental effects. To operationalize this principle, an effective way is to enforce the model to align its predictions under diverse conditional prompts—each simulating a distinct environmental distribution (e.g., randomly marking bounding box). Then, to learn causal factors, we get:
\begin{theorem}
    Let $\mathcal{D}$ denote the dataset for task $\tau$, and $\{ \mathcal{D}_i \}_{i=1}^{N_t}$ represent $N_t$ perturbed versions of $\mathcal{D}$ generated by random prompt annotations with the same samples $X$, i.e., $\mathcal{D}_i=\{X, P_i\}$. Let $f_\theta$ and $f_\phi^*$ denote the prompt optimizer and the optimal model for $\tau$. If for any $\mathcal{D}_i$ and $f$ we have $\mathcal{L}(f_\phi^*;X,f_\theta(P_i))\approx \varepsilon $ where $\varepsilon\ge 0$, then $f_\theta(P_i)$ is believed to be causal prompts with only causal factors. 
\end{theorem}
It shows that by making prompts generated under different conditions to enforce segmentation consistency on identical samples, the model is compelled to retain invariant causal factors without external samples. Further, according to \cite{treisman1980feature}, when the model achieves near-optimal performance (with $\varepsilon$ approaching 0), it can be inferred that the model has captured the most essential semantics required for segmentation. This suggests that the causal factors extracted by the model are largely complete in representing the semantic labels. See  \textbf{Appendix \ref{sec:app_proof}} for proofs.
Notably, Figure \ref{fig:prompt_bias_entity} shows that the same prompt in an OVMS sample may affect different entities differently. While causal multi-distribution consistency helps extract task-relevant causal factors, we also need to link these factors to specific entities. This ensures that each entity is segmented based on the most related factors. The detailed solution is provided in Section \ref{sec:4}.

\begin{figure*}[t]
    \centering
    \includegraphics[width=0.9\linewidth]{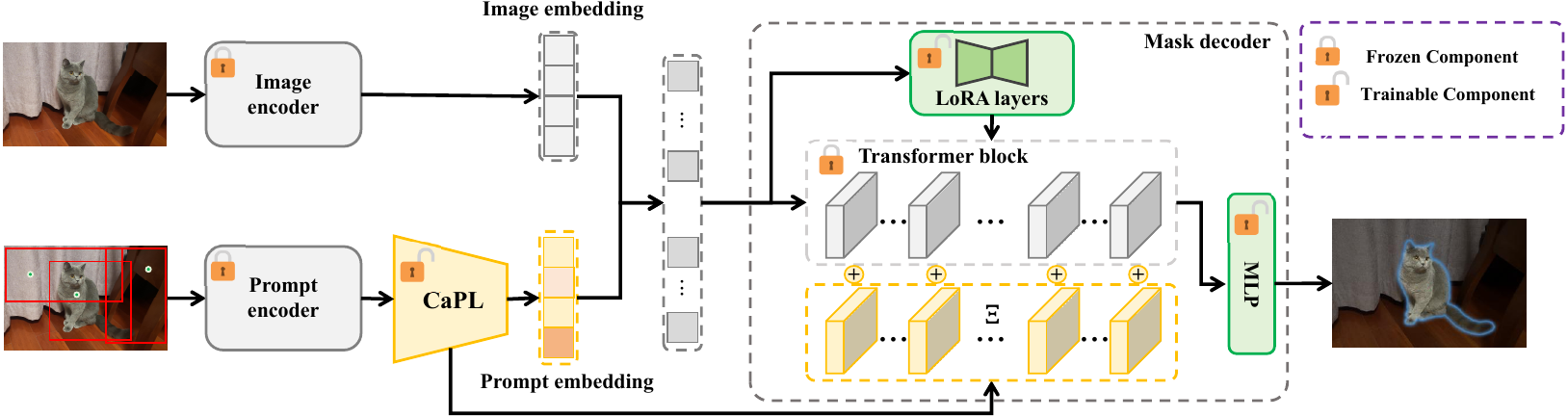}
    \vspace{-0.1in}
    \caption{The framework of CPC-SAM. The \textcolor[rgb]{0.6,0.5,0.0}{yellow box} represents CaPL and the \textcolor[rgb]{0,0.5,0}{green box} and \textcolor[rgb]{0.5,0.5,0.5}{gray box} are components for SAM. }
    \label{fig:mecasam}
    \vspace{-0.1in}
\end{figure*}

\section{Method}
\label{sec:4}

Based on the above analysis, we propose causal prompt calibration guided SAM (CPC-SAM) for accurate OVMS (\textbf{Figure \ref{fig:mecasam}}). It consists of the causal prompt learner (CaPL, $f_\theta$) for prompt calibration and SAM for segmentation ($f_\phi$). 
Specifically, we first generate multiple prompts for the same sample using random annotations to simulate diverse distributions. These prompts are then reweighted in CaPL to derive causal prompts by enforcing causal multi-distribution consistency. Given that the impact of task-relevant causal factors may vary across different entities (demonstrated in Figure \ref{fig:prompt_bias_entity}), CaPL consists of a task causal module and an entity causal module and performs calibration at both modules.  
To enhance reweighting accuracy, CaPL is optimized by minimizing the cumulative segmentation loss across different prompts, ensuring that the prompts capture only the causal factors relevant to the labels.  
To achieve precise OVMS on SAM, we introduce a bi-level optimization strategy that jointly optimizes CaPL and SAM. 
The overall objective is shown in \textbf{Subsection \ref{sec:method_overall}}.

\subsection{Prompt Calibration by CaPL}
\label{sec:method_inner}

In this subsection, we introduce the details of prompt calibration by CaPL.
Specifically, to simulate diverse distributions, we first generate multiple prompts for each sample using random annotations. Formally, given the data set $\mathcal{D}$ with $N$ samples, we perform $N_t$ interactions ($N_t=2$ in this paper), e.g., randomly annotate bounding boxes, and obtain $N_t$ groups of samples, denoted as $\mathcal{D}_i=\{x_{i,j},p_{i,j}\}_{j=1}^{N}$. Each group of samples $x_{i,j}$ is exactly the same except for the prompt $p_{i,j}$. Then, we input these groups of prompts into CaPL for reweighting. 
Considering the impact of prompt bias varies across entities, the reweighting is performed in two stages, including: (i) a task causal module, which reweights the prompt embeddings using multi-distribution consistency to extract causal factors, and (ii) an entity causal module, which reweights the image-to-token attention matrix using sparsity intervention to identify the relationships between causal factors and different entities. These two combine to form the causal prompt.

\noindent\textbf{Task Causal Module} aims to reweight the prompt embeddings to make it contain only task-relevant causal factors.
As demonstrated in \textbf{Theorem 1}, causal factors can be effectively extracted by enforcing consistency across multiple datasets with different distributions, i.e., the segmentation effect of the same model based on different groups of prompts is consistent. 
% This causal consistency allows us to reweight the prompt. 
Thus, to ensure the accurate output of this module, we propose to constrain it with causal multi-distribution consistency.
Then, given the $N_t$ datasets, we design the following loss function:
\begin{equation}\label{eq:task_causal}
\resizebox{0.9\linewidth}{!}{$
\begin{array}{l}
    \mathcal{L}_{CaPL}^{task}(f_\theta,f_\phi;\mathcal{D})=\sum_{i=1}^{N_t} \mathcal{L}_{SAM}(f_\theta,f_\phi;\mathcal{D}_i)\\[5pt]
    \text{s.t. }\scalebox{0.85}{$\mathcal{L}_{SAM}(f_\theta,f_\phi;\mathcal{D}_i)= \sum_{(x_{i,j},p_{i,j}) \in \mathcal{D}_i} y_{i,j} \log f_\phi(x_{i,j},f_\theta (p_{i,j}))$}
\end{array}$}
\end{equation}
where $f_\theta (p_{i,j})$ denotes the reweighted prompts via CaPL, $\mathcal{L}_{SAM}$ is the segmentation loss. 
By minimizing Eq.\ref{eq:task_causal}, we ensure that each set of prompts, weighted by CaPL, effectively guides SAM to produce optimal results, e.g., with losses approaching 0. This optimization process satisfies \textbf{Theorem 1}, ensuring the extraction of causal factors.

\noindent\textbf{Entity Causal Module} is also a calibration module. 
The backbone of SAM comprises multiple Transformer modules, each requiring multi-head attention computation, which involves calculating a weight matrix. This matrix represents the similarity between Queries and Keys, with its elements classified into three categories: (i) similarity between image patches and prompt patches, (ii) similarity between prompt patches and image patches, and (iii) similarity among prompt patches.
In the first module, prompts are constrained to extract generating factors of all entities. Consequently, (i) and (ii) reflect the response levels between different entities and their generative factors. In this module, CaPL aims to calibrate these relationships, ensuring that each entity exhibits a high response to its corresponding generative factor while maintaining minimal response to others.
It outputs a tensor matching the dimensions of the attention weight matrices across Transformer modules, where each multi-head attention matrix corresponds to a specific calibration matrix. Notably, at positions representing similarity among prompt patches, CaPL outputs zero. Furthermore, CaPL enforces sparsity in the output matrix to ensure a structured differentiation—allowing certain similarities to remain high while suppressing others to near zero—thus enhancing the disentanglement of generating factors.
Thus, we obtain the following loss function:
\begin{equation}\label{eq:entity_causal}
\resizebox{0.9\linewidth}{!}{$
    \scalebox{0.95}{$\mathcal{L}_{CaPL}^{entity}(f_\theta,f_\phi;\mathcal{D})=\sum_{e=1}^{N_e} \mathcal{L}_{SAM}(f_\theta,f_\phi;\Xi^e)  + \lambda_{\Xi} \| \Xi^e \|_1$}
    $}
\end{equation}
where $\lambda_{\Xi}$ is the weight hyperparameter, $N_{e}$ is the number of entities, $\Xi^e$ is the attention map of the $e$-th entity, $\mathcal{L}_{SAM}(f_\theta,f_\phi;\Xi^e)$ is the segmentation loss directly based on the attention matrix (see Appendix \ref{sec_app:implementation} for details). Notably, a key issue is whether the output of this module can accurately capture the similarity we aim to calibrate. To address this, we provide a detailed solution in \textbf{Subsection} \ref{sec:method_overall}.

\subsection{Fine-Tuning of SAM}
\label{sec:method_outer}

In this subsection, we introduce how to fine-tune SAM with CaPL. The role of CaPL is to provide reweighted prompts. Note that the optimization of SAM is implemented using LoRA \cite{hu2021lora,mi2024v}, aiming to prevent catastrophic forgetting and reduce resource consumption. The objective is:
\begin{equation}\label{eq:obj_sam}
\resizebox{0.9\linewidth}{!}{$
\begin{array}{l}
    \underset{\phi}{\min} \mathcal{L}_{SAM}(f_\theta,f_\phi;\mathcal{D})\\[5pt]
    \text{s.t. } \scalebox{0.85}{$\mathcal{L}_{SAM}(f_\theta,f_\phi;\mathcal{D})= \sum_{(x_{i,j},p_{i,j}) \in \mathcal{D}} y_{i,j} \log f_\phi(x_{i,j},f_\theta (p_{i,j}))$}
\end{array}
$}
\end{equation}
where $f_\theta$ denotes CaPL, and $\mathcal{L}_{SAM}$ can be this pixel-wise cross-entropy loss \cite{wang2021exploring,zhao2021contrastive} and also dice loss \cite{azad2023lossfunctionserasemantic}.

\subsection{Overall Optimization}
\label{sec:method_overall}
CPC-SAM utilizes bi-level optimization to integrate the above two interdependent optimization problems, thus achieving accurate OVMS. The overall objective is:
\begin{equation} \label{qwwweqwqwe}
\resizebox{0.9\linewidth}{!}{$
\begin{array}{*{20}{l}}
{\mathop {\min }\limits_{\theta ,\phi } {{\cal L}_{SAM}}({f_{\theta '(\theta )}},{f_\phi };{\cal D}) + {\lambda _{en}}{\cal L}_{CaPL}^{entity}({f_{\theta '(\theta )}},{f_\phi };{\cal D})}\\
{{\rm{s}}{\rm{.t}}{\rm{.}}\quad \theta '(\theta ) = \arg \mathop {\min }\limits_\theta  {{\cal L}_{CaPL}}({f_\theta },f_\phi ^*;{\cal D})}
\end{array}
$}
\end{equation}
where $\lambda_{en}$ is the weight of $\mathcal{L}_{CaPL}^{entity}$, which is set to 0.6 through parameter search (as illustrated in \textbf{Appendix \ref{sec_app:ex_ablation}}), ${{\cal L}_{CaPL}}( \cdot ) = {\cal L}_{CaPL}^{entity}( \cdot ) + {\cal L}_{CaPL}^{task}( \cdot )$, and ${\theta '(\theta )}$ denotes that $\theta '$ is a function of $\theta$.
During training, constraints are typically enforced through multiple gradient descent steps. If we explicitly write out the complete gradient descent process, we can express $\theta'$ as a function of $\theta$, thereby clearly revealing the relationship between parameter updates.

Note that the bi-level optimization also helps us constrain the output of the Entity Causal Module to accurately capture the similarity that needs to be calibrated. First, empirically, if the output of the Entity Causal Module is ideal, then it should be ``better'' than a non-ideal output. This notion of ``better'' can be measured by a smaller loss function value. For example, if $f_\theta^1$ is better than $f_\theta^2$, then $\mathcal{L}_{CaPL}^{entity}(f_\theta^1, f_\phi; \mathcal{D}) < \mathcal{L}_{CaPL}^{entity}(f_\theta^2, f_\phi; \mathcal{D})$. According to Eq.~(\ref{qwwweqwqwe}), we obtain a $\theta'$ under the constraint that minimizes $\mathcal{L}_{CaPL}^{entity}$. Then, in the main optimization objective of Eq.~(\ref{qwwweqwqwe}), by adjusting $\theta$, we change $\theta'$, which in turn modifies the value of $\mathcal{L}_{CaPL}^{entity}\bigl(f_{\theta'(\theta)}, f_\phi; \mathcal{D}\bigr)$, already at its minimum under the constraint, making it even smaller. Therefore, Eq.~(\ref{qwwweqwqwe}) uses bi-level optimization to constrain the output of the Entity Causal Module so that it accurately captures the similarity we aim to calibrate. In conclusion, Eq.~(\ref{qwwweqwqwe}) not only ensures the minimization of $\mathcal{L}_{CaPL}^{task}$, but also guarantees a further minimization of $\mathcal{L}_{CaPL}^{entity}$, thereby ultimately yielding an effective calibration.

\section{Empirical Evaluation}
\label{sec:6}

In this section, we first introduce the experimental setup (\textbf{Subsection \ref{sec:6.1}}). Next, we conduct comparison experiments on multiple scenarios to evaluate the performance of CPC-SAM (\textbf{Subsection \ref{sec:6.2}}). Finally, we perform ablation studies (\textbf{Subsection \ref{sec:6.3}}). 
All results reported are the averages of five independent runs NVIDIA A100 GPUs. More details and experiments are provided in \textbf{Appendices \ref{sec_app:dataset}-\ref{sec_app:experiment}}.

\begin{table*}
\centering
\caption{Performance comparison of general target OVMS (average Dice score (\%) with standard deviations) on CelebAMask-HQ, intelecaiCAR, and TikTok dance with \textbf{the standard setting}, i.e., trained with $N_b$ labeled examples. The best results are highlighted in \textbf{bold}.}
\vspace{-0.1in}
\label{tab:1_common}
    \resizebox{1\linewidth}{!}{
    \begin{tabular}{lcccccccccccc}
\toprule
    \multirow{2.5}{*}{Method} &\multicolumn{6}{c}{CelebAMask-HQ ($N_b=8$)} & \multicolumn{4}{c}{intelecaiCAR ($N_b=4$)}  & \multirow{2.5}{*}{TikTok dances ($N_b=8$)} & \multirow{2.5}{*}{Training Cost} \\
    \cmidrule(lr){2-7} \cmidrule(lr){8-11}  
    & Brow & Eye & Hair & Nose & Mouth & Overall & Body & Wheel & Window & Overall \\
\midrule
    DeepLabV3 \cite{chen2017rethinking} & 32.2 $\pm$ 0.6 & 54.4 $\pm$ 1.4 & 53.5 $\pm$ 1.7 & 62.6 $\pm$ 0.2 & 52.8 $\pm$ 0.2 & 52.9 $\pm$ 0.8 & 59.4 $\pm$ 0.8 & 64.3 $\pm$ 0.7 & 44.5 $\pm$ 1.6 & 57.1 $\pm$ 0.5 & 34.6 $\pm$ 0.4 & 4.5 \\
    SwinUnet \cite{cao2022swin} & 24.4 $\pm$ 1.3 & 30.0 $\pm$ 0.8 & 47.8 $\pm$ 1.2 & 45.1 $\pm$ 0.7 & 45.8 $\pm$ 0.9 & 38.7 $\pm$ 1.2 & 47.6 $\pm$ 4.3 & 51.8 $\pm$ 1.2 & 35.7 $\pm$ 1.4 & 44.1 $\pm$ 0.7 & 53.2 $\pm$ 0.2 &  4.1 \\
    RITM \cite{sofiiuk2022reviving} & 28.6 $\pm$ 1.5 & 24.0 $\pm$ 1.0 & 40.7 $\pm$ 1.1 & 40.2 $\pm$ 0.9 & 40.5 $\pm$ 0.9 & 33.7 $\pm$ 1.1 & 41.3 $\pm$ 2.1 & 48.5 $\pm$ 0.9 & 29.3 $\pm$ 1.4 & 40.5 $\pm$ 1.2 & 47.0 $\pm$ 0.7 & 3.2 \\
    ViTDet-H \cite{li2022exploring} & 25.5 $\pm$ 0.9 & 21.1 $\pm$ 1.1 & 38.4 $\pm$ 1.4 & 41.0 $\pm$ 1.3 & 38.9 $\pm$ 1.0 & 32.4 $\pm$ 0.8 & 39.0 $\pm$ 1.0 & 45.2 $\pm$ 0.7 & 27.0 $\pm$ 1.0 & 39.6 $\pm$ 0.9 & 44.8 $\pm$ 1.4 & 4.5 \\
\midrule
    HSNet \cite{zhang2022hsnet} & 40.6 $\pm$ 1.2 & 55.1 $\pm$ 0.4 & 72.2 $\pm$ 2.5 & 55.8 $\pm$ 1.8 & 62.1 $\pm$ 0.4 & 56.9 $\pm$ 0.6 & 67.8 $\pm$ 0.5 & 66.9 $\pm$ 0.1 & 61.9 $\pm$ 0.5 & 65.5 $\pm$ 0.3 & 52.8 $\pm$ 1.7 & 3.9  \\
    SSP \cite{fan2022self} & 43.6 $\pm$ 0.2 & 27.4 $\pm$ 0.8 & 72.9 $\pm$ 0.7 & 71.6 $\pm$ 1.1 & 66.7 $\pm$ 0.4 & 57.1 $\pm$ 1.1 & 73.6 $\pm$ 0.1 & 75.8 $\pm$ 0.2 & 58.1 $\pm$ 1.4 & 67.5 $\pm$ 0.8 & 73.5 $\pm$ 0.6 & 5.1 \\
\midrule
    Vanilla SAM \cite{sam} & 17.6 $\pm$ 1.5 & 28.4 $\pm$ 0.5 & 40.2 $\pm$ 0.8 & 40.9 $\pm$ 0.8 & 23.7 $\pm$ 3.2 & 30.3 $\pm$ 0.7 & 37.4 $\pm$ 1.8 & 38.6 $\pm$ 0.7 & 19.6 $\pm$ 0.7 & 31.8 $\pm$ 1.1 & 23.1 $\pm$ 0.8 & 5.1 \\
    Med-SA \cite{wu2023medical} & 41.2 $\pm$ 0.4 & 62.8 $\pm$ 1.2 & 80.5 $\pm$ 0.2 & 72.1 $\pm$ 0.4 & 70.8 $\pm$ 0.3 & 65.4 $\pm$ 0.6 & 82.1 $\pm$ 0.7 & 74.9 $\pm$ 0.2 & 62.8 $\pm$ 2.1 & 73.2 $\pm$ 1.5 & 76.9 $\pm$ 0.3 & 4.9  \\
    SAMed \cite{tancik2020fourier} & 36.4 $\pm$ 1.2 & 63.9 $\pm$ 0.2 & 79.1 $\pm$ 0.4 & 67.8 $\pm$ 0.9 & 64.7 $\pm$ 0.8 & 62.3 $\pm$ 0.2 & 82.7 $\pm$ 0.4 & 68.8 $\pm$ 0.5 & 60.9 $\pm$ 2.1 & 70.8 $\pm$ 1.2 & 78.4 $\pm$ 0.7 & 5.6 \\
    BLO-SAM \cite{zhangblo} & 41.8 $\pm$ 0.4 & 67.6 $\pm$ 0.7 & 81.6 $\pm$ 0.4 & 72.6 $\pm$ 0.3 & 69.2 $\pm$ 0.5 & 65.1 $\pm$ 1.3 & 83.3 $\pm$ 0.5 & 76.4 $\pm$ 0.4 & 63.3 $\pm$ 0.2 & 75.0 $\pm$ 0.4 & 82.1 $\pm$ 0.3 & 7.0  \\
    OVSAM \cite{yuan2024open} & 45.7 $\pm$ 1.0 & 65.7 $\pm$ 0.8 & 85.7 $\pm$ 0.6 & 67.2 $\pm$ 0.7 & 71.9 $\pm$ 1.2 & 67.5 $\pm$ 0.8 & 83.1 $\pm$ 1.1 & 76.3 $\pm$ 0.4 & 65.5 $\pm$ 1.5 & 78.3 $\pm$ 0.8 & 84.9 $\pm$ 0.5 & 9.2 \\
\midrule
    ZS3Net \cite{bucher2019zero} & 24.0 $\pm$ 1.1 & 34.8 $\pm$ 0.6 & 49.6 $\pm$ 0.7 & 46.8 $\pm$ 0.5 & 33.9 $\pm$ 1.1 & 38.4 $\pm$ 0.6 & 42.1 $\pm$ 0.8 & 48.2 $\pm$ 0.3 & 23.7 $\pm$ 1.1 & 38.4 $\pm$ 0.4 & 33.9 $\pm$ 0.6 & 4.7 \\
    LSeg \cite{li2022language} & 37.5 $\pm$ 0.9 & 49.5 $\pm$ 0.9 & 62.4 $\pm$ 0.3 & 58.1 $\pm$ 0.6 & 45.6 $\pm$ 0.7 & 51.4 $\pm$ 0.4 & 54.8 $\pm$ 0.6 & 57.9 $\pm$ 0.9 & 40.5 $\pm$ 0.8 & 53.7 $\pm$ 1.1 & 44.5 $\pm$ 1.3 & 4.5 \\
    ZegFormer \cite{ding2022decoupling} & 29.7 $\pm$ 1.2 & 41.2 $\pm$ 1.1 & 54.7 $\pm$ 0.9 & 55.1 $\pm$ 1.2 & 32.4 $\pm$ 0.6 & 44.8 $\pm$ 0.7 & 51.6 $\pm$ 0.9 & 53.2 $\pm$ 0.4 & 30.7 $\pm$ 0.6 & 45.7 $\pm$ 0.5 & 37.6 $\pm$ 0.4 & 6.0 \\
    OpenSeg \cite{ghiasi2022scaling} & 40.7 $\pm$ 1.4 & 59.4 $\pm$ 0.8 & 73.8 $\pm$ 0.4 & 65.1 $\pm$ 0.7 & 42.8 $\pm$ 0.8 & 52.6 $\pm$ 0.4 & 69.5 $\pm$ 1.3 & 73.7 $\pm$ 0.6 & 52.7 $\pm$ 0.9 & 60.4 $\pm$ 0.7 & 55.8 $\pm$ 0.5 & 8.4 \\
    OVSeg \cite{liang2023open} & 44.9 $\pm$ 1.0 & 52.9 $\pm$ 1.2 & 75.4 $\pm$ 0.8 & 67.4 $\pm$ 0.5 & 67.6 $\pm$ 0.5 & 63.1 $\pm$ 0.5 & 75.1 $\pm$ 0.5 & 77.4 $\pm$ 1.2 & 63.7 $\pm$ 0.4 & 73.6 $\pm$ 0.5 & 79.8 $\pm$ 0.7 & 7.6 \\
    CPC-SAM & \textbf{50.2 $\pm$ 0.8} & \textbf{71.2 $\pm$ 1.1} & \textbf{88.0 $\pm$ 0.9} & \textbf{73.0 $\pm$ 1.2} & \textbf{72.1$\pm$ 1.3} & \textbf{70.4 $\pm$ 0.8} & \textbf{84.1 $\pm$ 1.1} & \textbf{77.6 $\pm$ 0.9} & \textbf{64.5 $\pm$ 1.2} & \textbf{80.2 $\pm$ 0.7} & \textbf{88.3 $\pm$ 0.3} & 4.7 \\
\bottomrule
\end{tabular}}
\end{table*}

\begin{table*}[t]
\centering
\caption{Performance comparison of general target OVMS (average Dice score (\%) with standard deviations) on CelebAMask-HQ, intelecaiCAR, and TikTok dance with \textbf{few-shot setting}. $N_b$ denotes the number of labeled examples. The best results are highlighted in \textbf{bold}.}
\vspace{-0.1in}
\label{tab:1_few-shot}
    \resizebox{1\linewidth}{!}{
    \begin{tabular}{lcccccccccccc}
\toprule
    \multirow{2.5}{*}{Method} &\multicolumn{6}{c}{CelebAMask-HQ ($N_b=4$)} & \multicolumn{4}{c}{intelecaiCAR ($N_b=2$)}  & \multirow{2.5}{*}{TikTok dances ($N_b=4$)} & \multirow{2.5}{*}{Training Cost} \\
    \cmidrule(lr){2-7} \cmidrule(lr){8-11}  
    & Brow & Eye & Hair & Nose & Mouth & Overall & Body & Wheel & Window & Overall \\
\midrule
    DeepLabV3 \cite{chen2017rethinking} & 29.2 $\pm$ 0.4 & 50.6 $\pm$ 1.8 & 49.4 $\pm$ 4.2 & 52.8 $\pm$ 2.4 & 43.2 $\pm$ 3.7 & 45.1 $\pm$ 1.6 & 54.2 $\pm$ 4.3 & 51.6 $\pm$ 2.7 & 23.8 $\pm$ 3.8 & 42.8 $\pm$ 3.9 & 29.7 $\pm$ 1.6 & 2.2 \\
    SwinUnet \cite{cao2022swin} & 18.9 $\pm$ 1.1 & 18.4 $\pm$ 1.6 & 43.8 $\pm$ 1.1 & 42.1 $\pm$ 1.3 & 38.6 $\pm$ 0.7 & 32.3 $\pm$ 0.9 & 19.6 $\pm$ 5.4 & 29.6 $\pm$ 3.7 & 34.2 $\pm$ 6.8 & 27.8 $\pm$ 0.9 & 27.3 $\pm$ 2.4 & 3.7  \\
    RITM \cite{sofiiuk2022reviving} & 22.0 $\pm$ 0.7 & 22.6 $\pm$ 1.7 & 27.5 $\pm$ 1.8 & 38.1 $\pm$ 1.6 & 34.9 $\pm$ 1.4 & 31.7 $\pm$ 0.5 & 35.5 $\pm$ 4.6 & 44.3 $\pm$ 3.4 & 24.2 $\pm$ 5.4 & 36.4 $\pm$ 1.5 & 35.6 $\pm$ 2.1 & 3.0  \\
    ViTDet-H \cite{li2022exploring} & 20.1 $\pm$ 0.8 & 19.8 $\pm$ 1.4 & 35.6 $\pm$ 2.7 & 37.7 $\pm$ 1.4 & 36.8 $\pm$ 1.9 & 30.1 $\pm$ 1.2 & 35.7 $\pm$ 4.5 & 41.8 $\pm$ 2.6 & 22.7 $\pm$ 6.2 & 34.6 $\pm$ 1.4 & 36.4 $\pm$ 1.5 & 4.1  \\
\midrule
    HSNet \cite{zhang2022hsnet} & 27.5 $\pm$ 1.8 & 38.4 $\pm$ 0.8 & 56.2 $\pm$ 1.6 & 57.6 $\pm$ 1.3 & 54.3 $\pm$ 0.7 & 46.8 $\pm$ 0.6 & 65.8 $\pm$ 1.4 & 28.7 $\pm$ 0.5 & 50.8 $\pm$ 0.7 & 48.4 $\pm$ 0.8 & 46.8 $\pm$ 2.3 & 3.3  \\
    SSP \cite{fan2022self} & 36.8 $\pm$ 0.4 & 29.5 $\pm$ 0.8 & 68.9 $\pm$ 0.1 & 67.8 $\pm$ 0.4 & 62.8 $\pm$ 0.6 & 53.1 $\pm$ 0.5 & 58.2 $\pm$ 0.6 & 73.9 $\pm$ 0.8 & 49.1 $\pm$ 0.3 & 60.4 $\pm$ 0.5 & 56.1 $\pm$ 0.5 & 3.9  \\
\midrule
    Vanilla SAM \cite{sam} & 17.2 $\pm$ 1.2 & 26.3 $\pm$ 0.7 & 40.2 $\pm$ 0.6 & 42.1 $\pm$ 0.5 & 20.4 $\pm$ 3.6 & 29.2 $\pm$ 0.7 & 36.4 $\pm$ 1.8 & 37.2 $\pm$ 0.5 & 18.9 $\pm$ 0.7 & 30.8 $\pm$ 1.2 & 19.3 $\pm$ 0.7 & 4.0 \\
    Med-SA \cite{wu2023medical} & 30.9 $\pm$ 1.1 & 57.6 $\pm$ 0.2 & 72.5 $\pm$ 0.8 & 67.9 $\pm$ 0.4 & 61.3 $\pm$ 0.6 & 58.6 $\pm$ 0.5 & 76.8 $\pm$ 0.7 & 68.4 $\pm$ 0.2 & 47.6 $\pm$ 3.1 & 64.2 $\pm$ 2.3 & 54.3 $\pm$ 0.4 & 4.2  \\
    SAMed \cite{tancik2020fourier} & 24.8 $\pm$ 0.7 & 51.4 $\pm$ 1.6 & 73.9 $\pm$ 0.8 & 63.4 $\pm$ 1.3 & 59.4 $\pm$ 0.1 & 54.5 $\pm$ 1.1 & 75.9 $\pm$ 0.8 & 62.4 $\pm$ 1.1 & 34.2 $\pm$ 1.3 & 57.5 $\pm$ 0.8 & 60.9 $\pm$ 0.8 & 4.5  \\
    BLO-SAM \cite{zhangblo} & 36.8 $\pm$ 0.4 & 61.8 $\pm$ 0.7 & 79.3 $\pm$ 0.6 & 70.9 $\pm$ 0.7 & 64.2 $\pm$ 0.8 & 62.6 $\pm$ 0.6 & 79.6 $\pm$ 0.2 & 71.8 $\pm$ 0.8 & 52.7 $\pm$ 0.7 & 68.1 $\pm$ 0.5 & 73.8 $\pm$ 0.7 & 5.7  \\
    OVSAM \cite{yuan2024open} & 40.8 $\pm$ 1.0 & 61.4 $\pm$ 1.3 & 81.4 $\pm$ 1.1 & 66.8 $\pm$ 0.5 & 67.7 $\pm$ 1.2 & 65.7 $\pm$ 0.4 & 65.3 $\pm$ 1.3 & 72.6 $\pm$ 0.4 & 52.5 $\pm$ 1.6 & 70.4 $\pm$ 0.7 & 74.5 $\pm$ 0.9 & 7.4  \\
\midrule
    ZS3Net \cite{bucher2019zero} & 19.2 $\pm$ 0.7 & 28.6 $\pm$ 0.8 & 43.8 $\pm$ 0.7 & 45.9 $\pm$ 0.6 & 23.7 $\pm$ 0.4 & 33.7 $\pm$ 0.8 & 39.1 $\pm$ 0.7 & 40.5 $\pm$ 0.6 & 22.9 $\pm$ 1.2 & 33.7 $\pm$ 1.4 & 20.6 $\pm$ 0.5 & 4.2 \\
    LSeg \cite{li2022language} & 27.0 $\pm$ 1.1 & 35.4 $\pm$ 0.5 & 53.7 $\pm$ 0.4 & 50.5 $\pm$ 0.9 & 33.6 $\pm$ 0.7 & 42.8 $\pm$ 0.5 & 45.1 $\pm$ 0.4 & 49.3 $\pm$ 0.8 & 25.4 $\pm$ 0.9 & 41.8 $\pm$ 0.8 & 30.4 $\pm$ 0.6 & 3.5 \\
    ZegFormer \cite{ding2022decoupling} & 22.7 $\pm$ 0.9 & 32.7 $\pm$ 0.9 & 47.4 $\pm$ 1..3 & 49.6 $\pm$ 0.8 & 23.9 $\pm$ 0.4 & 36.1 $\pm$ 0.7 & 42.8 $\pm$ 1.1 & 40.9 $\pm$ 0.7 & 25.8 $\pm$ 0.5 & 37.4 $\pm$ 1.1 & 25.2 $\pm$ 1.2 & 4.9 \\
    OpenSeg \cite{ghiasi2022scaling} & 36.5 $\pm$ 0.7 & 57.2 $\pm$ 1.1 & 67.1 $\pm$ 0.5 & 64.2 $\pm$ 1.2 & 36.4 $\pm$ 1.1 & 50.3 $\pm$ 0.9 & 66.7 $\pm$ 0.5 & 68.4 $\pm$ 0.6 & 45.1 $\pm$ 1.1 & 55.9 $\pm$ 0.5 & 49.6 $\pm$ 1.1 & 5.9 \\
    OVSeg \cite{liang2023open} & 41.0 $\pm$ 1.2 & 59.3 $\pm$ 1.1 & 73.0 $\pm$ 0.8 & 66.9 $\pm$ 1.2 & 62.1 $\pm$ 1.0 & 57.2 $\pm$ 1.5 & 72.8 $\pm$ 0.9 & \textbf{72.0 $\pm$ 0.8} & 59.6 $\pm$ 1.1 & 64.2 $\pm$ 0.9 & 70.0 $\pm$ 1.0 & 7.0 \\
    CPC-SAM & \textbf{46.9 $\pm$ 0.7} & \textbf{65.9 $\pm$ 1.1} & \textbf{83.2 $\pm$ 0.7} & \textbf{68.3 $\pm$ 0.9} & \textbf{68.0 $\pm$ 1.1} & \textbf{67.8 $\pm$ 0.9} & \textbf{81.8 $\pm$ 0.6} & 71.9 $\pm$ 0.9 & \textbf{60.0 $\pm$ 1.0} & \textbf{74.9 $\pm$ 1.2} & \textbf{80.0 $\pm$ 0.5} & 3.7 \\
\bottomrule
\end{tabular}}
\end{table*}

\begin{table}
\vspace{-0.1in}
\centering
\caption{Performance comparison of medical OVMS (average Dice score (\%) with standard deviations) on BBBC038v1, Kvasir-SEG, and JSRT. "-" means the result is not recorded, and ``OOD'' means being trained on Kvasir-SEG and then tested on JSRT.}
\vspace{-0.1in}
\label{tab:2}
    \resizebox{1\linewidth}{!}{
    \begin{tabular}{lccccccc}
\toprule
    \multirow{2.5}{*}{Method} & \multicolumn{2}{c}{BBBC038v1} & \multicolumn{2}{c}{Kvasir-SEG}  & \multicolumn{2}{c}{JSRT} & \multirow{2.5}{*}{OOD} \\
    \cmidrule(lr){2-3} \cmidrule(lr){4-5} \cmidrule(lr){6-7} 
    & $N_b=8$ & $N_b=4$ & $N_b=8$ & $N_b=4$ & $N_b=4$ & $N_b=2$ \\
\midrule
    DeepLabV3 & 46.3 $\pm$ 0.6 & 31.0 $\pm$ 1.1 & 34.6 $\pm$ 0.7 & 26.8 $\pm$ 0.4 & 74.1 $\pm$ 0.9 & 58.6 $\pm$ 0.8 & 14.4 \\
    SwinUnet & 42.1 $\pm$ 0.7 & 38.1 $\pm$ 0.9 & 35.2 $\pm$ 0.8 & 34.6 $\pm$ 0.2 & 74.8 $\pm$ 0.4 & 60.5 $\pm$ 0.7 & - \\
    % RITM &  &  &  &  &  &  &  \\
    % ViTDet-H &  &  &  &  &  &  &  \\
\midrule
    HSNet & - & - & 31.8 $\pm$ 0.6 & 26.7 $\pm$ 0.5 & 82.1 $\pm$ 0.7 & 80.9 $\pm$ 0.2 & - \\
    SSP & 43.1 $\pm$ 0.7 & 38.0 $\pm$ 0.7 & 25.6 $\pm$ 0.9 & 23.4 $\pm$ 0.3 & 86.7 $\pm$ 0.8 & 79.4 $\pm$ 0.1 & 15.1 \\
\midrule
    Vanilla SAM & 42.6 $\pm$ 1.0 & 34.2 $\pm$ 1.3 & 13.8 $\pm$ 0.4 & 10.3 $\pm$ 0.9 & 29.8 $\pm$ 0.5 & 27.2 $\pm$ 0.7 & 10.4 \\
    Med-SA & - & - & 56.4 $\pm$ 0.7 & 30.1 $\pm$ 0.5 & 87.6 $\pm$ 0.4 & 79.8 $\pm$ 0.3 & - \\
    SAMed & 50.1 $\pm$ 0.9 & 46.5 $\pm$ 1.0 & 54.1 $\pm$ 0.4 & 39.8 $\pm$ 0.2 & 88.8 $\pm$ 0.6 & 82.9 $\pm$ 0.2 & 15.5 \\
    BLO-SAM & 76.4 $\pm$ 0.9 & 68.6 $\pm$ 0.7 & 59.2 $\pm$ 0.7 & 56.4 $\pm$ 0.8 & 90.5 $\pm$ 0.4 & 82.6 $\pm$ 0.9 & 22.9 \\
    OVSAM & 72.1 $\pm$ 0.7 & 62.0 $\pm$ 0.7 & 58.7 $\pm$ 1.1 & 55.2 $\pm$ 0.9 & 91.2 $\pm$ 0.7 & 84.6 $\pm$ 0.9 & 22.4 \\
\midrule
    ZS3Net & - & - & 16.2 $\pm$ 0.7 & 13.1 $\pm$ 0.8 & - & - & - \\
    LSeg & - & - & 18.9 $\pm$ 1.0 & 16.6 $\pm$ 1.1 & - & - & - \\
    OpenSeg & 69.8 $\pm$ 0.7 & 60.2 $\pm$ 0.8 & 48.8 $\pm$ 1.1 & 45.1 $\pm$ 1.4 & - & - & - \\
    OVSeg & 70.3 $\pm$ 0.5 & 62.1 $\pm$ 0.4 & 57.6 $\pm$ 1.2 & 49.1 $\pm$ 0.7 & 89.9 $\pm$ 0.6 & 81.5 $\pm$ 0.7 & 21.4 \\
    CPC-SAM & \textbf{82.2 $\pm$ 1.0} & \textbf{75.5 $\pm$ 0.9} & \textbf{64.4 $\pm$ 0.7} & \textbf{59.9 $\pm$ 1.0} & \textbf{94.9 $\pm$ 0.7} & \textbf{87.1 $\pm$ 0.5} & \textbf{35.9} \\
\bottomrule
\end{tabular}}
\vspace{-0.15in}
\end{table}

\begin{figure*}
\begin{subfigure}{0.3\linewidth}
        \centering
        \includegraphics[width=\textwidth]{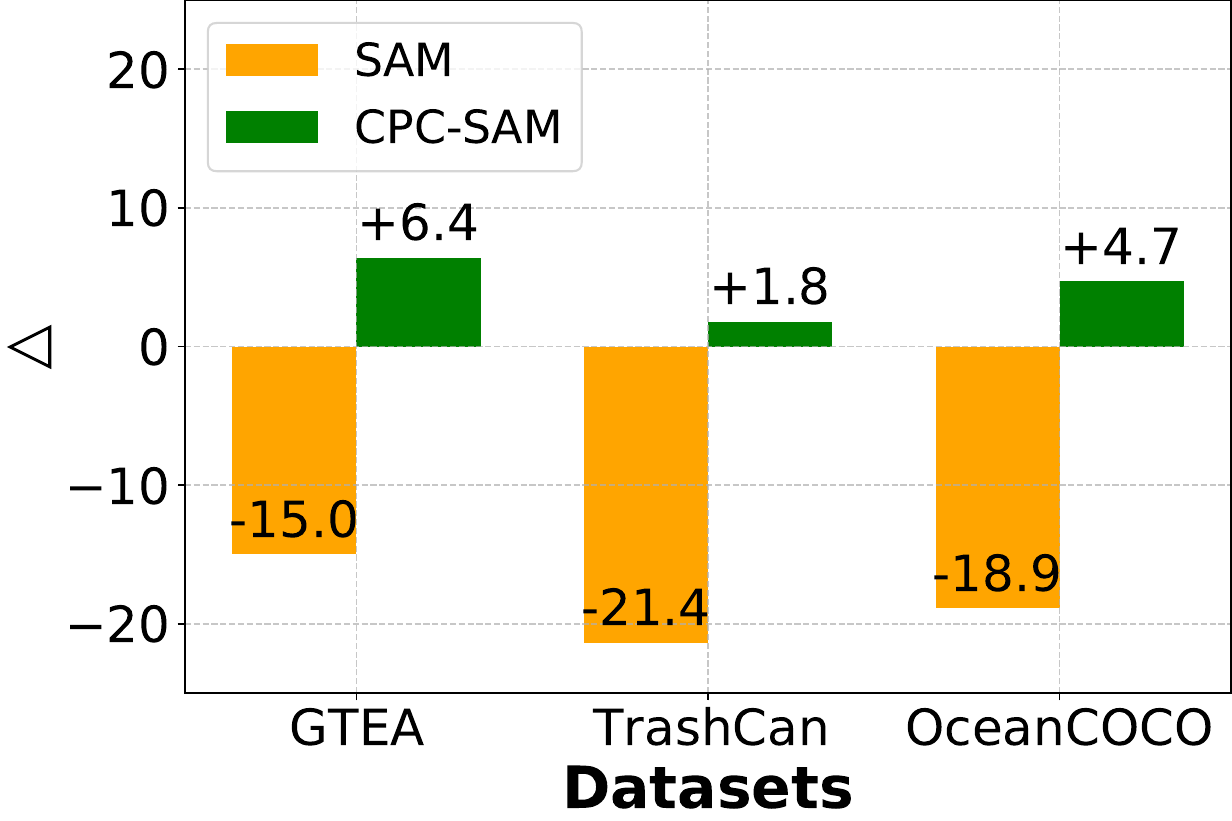}
        \caption{SAM, CPC-SAM vs. RITM}
        \label{fig:ex_3_1}
    \end{subfigure}
    \hfill
    \begin{subfigure}{0.66\linewidth}
        \centering
        \includegraphics[width=\textwidth]{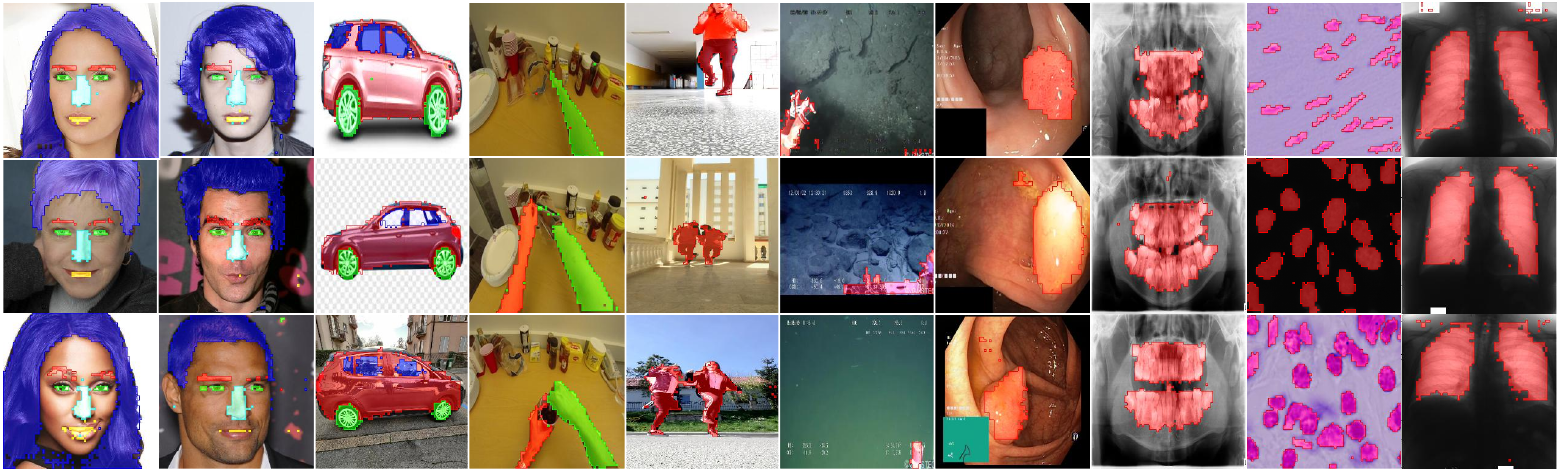}
        \caption{OVMS visualization} 
        \label{fig:ex_3_2}
    \end{subfigure}
    \vspace{-0.1in}
    \caption{Performance comparison of complex scene and activity segmentation and visualization. (a) shows the performance gap ($\bigtriangleup $) of SAM and CPC-SAM vs. RITA on three datasets. (b) shows the OVMS visualization of CPC-SAM, with more results in \textbf{Appendix \ref{sec_app:experiment}}.}
    \label{fig:ex_3}
\end{figure*}

\begin{figure}
% \vspace{-0.1in}
    \begin{minipage}[t]{0.48\columnwidth}
        \centering
        \includegraphics[width=\linewidth]{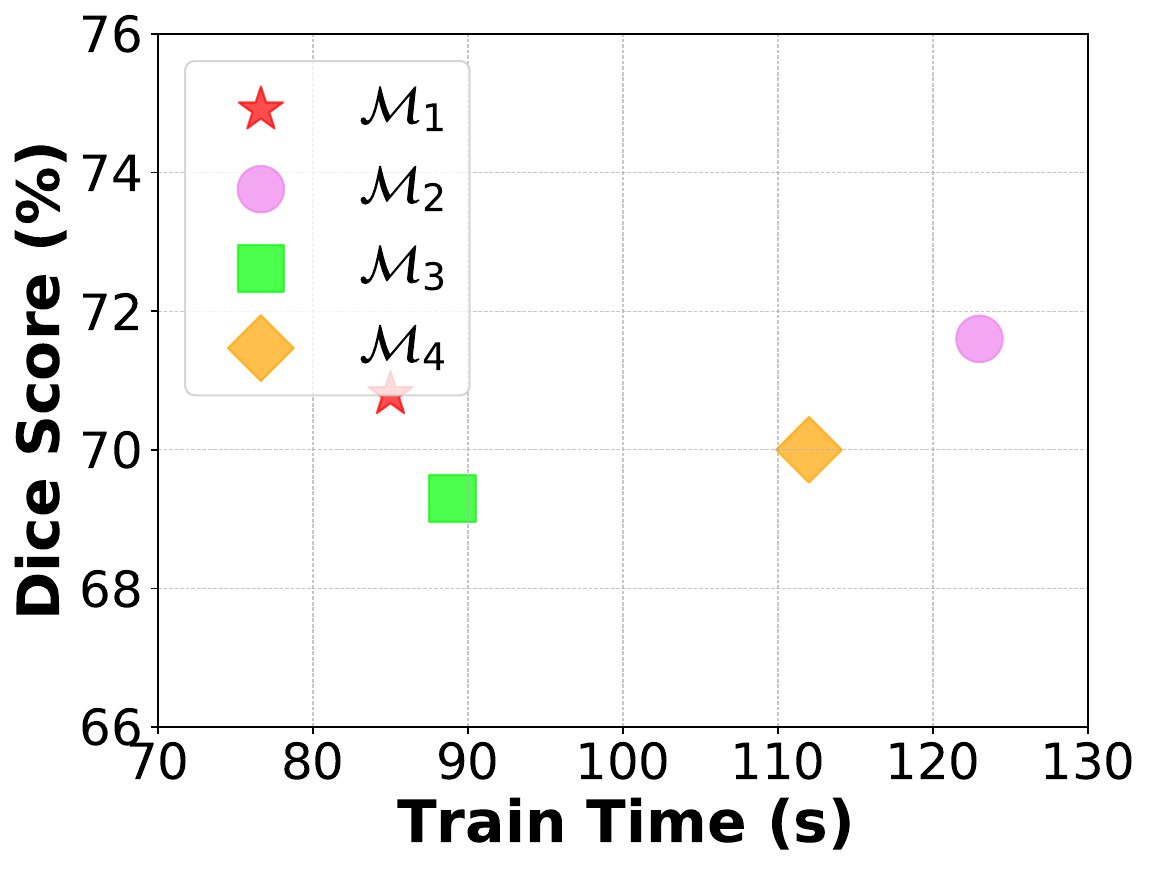}
        \vspace{-0.25in}
        \caption{Ablation study on prompt optimization methods.}
        \vspace{-0.1in}
        \label{fig:abla_prompt}
    \end{minipage}
    \hfill
    \begin{minipage}[t]{0.48\columnwidth}
        \centering
        \includegraphics[width=\linewidth]{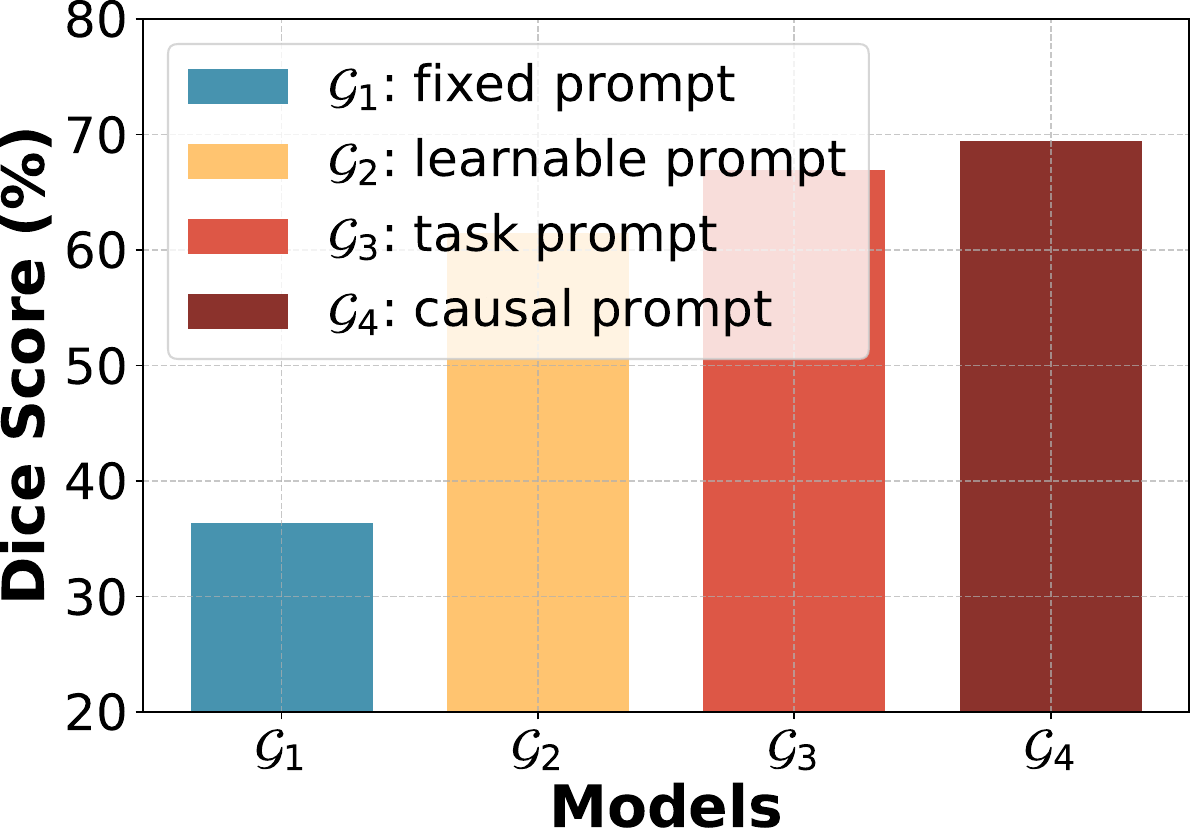}
        \vspace{-0.25in}
        \caption{Ablation study on different types of prompts.}
        \vspace{-0.1in}
        \label{fig:abla_1_toy}
    \end{minipage}
    % \vspace{-0.25in}
\end{figure}

\subsection{Experimental Setup}
\label{sec:6.1}

\noindent\textbf{Datasets}
We evaluate CPC-SAM across a diverse set of semantic segmentation tasks on nine benchmark datasets following \cite{sam,zhangblo,yuan2024open}. These datasets include (i) general target segmentation: CelebAMask-HQ \cite{lee2020maskgan}, intelecaiCAR \cite{david2020carparts}, and TikTok dances \cite{roman2023humantiktok} for human facial components, car, and human body segmentation, (ii) medical segmentation: BBBC038v1 \cite{caicedo2019nucleus}, Kvasir-SEG \cite{jha2020kvasir}, and JSRT dataset \cite{shiraishi2000development} for medical cell, gastrointestinal, and lung segmentation, and (iii) complex scene and activity segmentation: GTEA \cite{fathi2011learning}, TrashCan \cite{hong2020trashcan}, and self-made OceanCOCO for daily activities, underwater trash, and deepsea scene segmentation. Note that GTEA and TrashCan are the two worst-performing datasets of SAM \cite{sam}, while OceanCOCO consists of the benchmark COCO \cite{lin2014microsoft} and self-collected deep-sea samples of previously unseen classes. 
% The datasets span multiple domains and segmentation difficulties, providing a comprehensive evaluation of CPC-SAM's applicability. 
\emph{For OVMS, each dataset is split into annotated base and unannotated target classes in a $2:1$ ratio, with training based on the base class annotations.} More details are provided in \textbf{Appendix \ref{sec_app:dataset}}.

\noindent\textbf{Implementation Details} 

CPC-SAM consists of CaPL and SAM. During training, these models are alternately optimized. For SAM, we use the ViT-B model as the image encoder, while the causal prompt learner employs a simple 2-layer MLP to calibrate prompts. 
In our approach, each attention layer of SAM can be divided into three types: the prompt’s self-attention, the cross-attention from the prompt to the image, and the cross-attention from the image to the prompt. We only apply an L1 constraint on the attention map of the prompt’s self-attention and the cross-attention from the image to the prompt.
The LoRA layers and unfrozen components are optimized using the Adam optimizer, with an initial learning rate of $5e-3$ and weight decay of $0.1$ following \cite{sam,tancik2020fourier}. The training spanned $100$ epochs, and the optimal checkpoint is selected based on OVMS performance. See \textbf{Appendix \ref{sec_app:implementation}} for more details.

\subsection{Results and Analysis}
\label{sec:6.2}
In this subsection, we provide the results and analyses of the comparison experiments, evaluating the effectiveness of CPC-SAM. More results are provided in \textbf{Appendices \ref{sec_app:dataset}-\ref{sec_app:experiment}}.

% \vspace{-0.25in}
\noindent\textbf{Results on General Target Segmentation}

In OVMS tasks for general targets, we establish two training settings: the standard setting with batch size $N_b$ and a few-shot setting with batch size $N_b/2$. 
Here, $N_b$ represents the typical number of training samples per batch, e.g., $N_b=8$ for CelebAMask-HQ.
% Here, $N$ represents the typical number of training samples per batch, e.g., $N=8$ for CelebAMask, $N=4$ for intelecaiCAR, and $N=8$ for TikTok dances. 
We measure each model’s performance in both settings using the Dice score and Training Cost to evaluate their effectiveness under normal and data-limited conditions. 
The results are shown in \textbf{Table \ref{tab:1_common}} (standard) and \textbf{Table \ref{tab:1_few-shot}} (few-shot), respectively.
We observe that: (i) CPC-SAM exhibits robust capability, achieving accurate OVMS with limited samples, e.g., surpasses 70\% accuracy on CelebAMask for the first time. 
(ii) CPC-SAM consistently outperforms previous methods, including OVSAM from SAM-based baselines and OVSeg from OVMS-specific baselines, with an average improvement of 3.9\%, expanding to 5.1\% in the few-shot setting. (iii) CPC-SAM delivers an excellent trade-off between high Dice scores and low training costs. These results highlight the advantages of CPC-SAM.

\noindent\textbf{Results on Medical Segmentation}

Given the complexity of medical imaging, we conduct evaluation for medical segmentation. 
To assess model robustness, we use datasets with varying target scales, i.e., BBBC038v1, Kvasir-SEG, and JSRT, under standard and few-shot settings. We also evaluate the performance of JSRT after training on Kvasir-SEG to examine OOD generalization.
The results are shown in \textbf{Table \ref{tab:2}}. We observe that CPC-SAM outperforms the baselines: its performance exceeds the SOTA baseline by 5.5\% ($N_b=8$), while outperforming it by exceeding 9.8\% on average in few-shot settings ($N_b=4$); in the OOD setting, it exceeds the SOTA baseline by approximately 13\%. This demonstrates CPC-SAM's effectiveness in segmenting small targets and complex features.

\noindent\textbf{Results on Complex Scene and Activity Segmentation}
We also conduct experiments on complex scene and activity segmentation tasks. 
We record the Mean IoU of  SAM and CPC-SAM compared with the strongest single point segmenter RITM \cite{sofiiuk2022reviving} on GTEA, TrashCan, and OceanCOCO following \cite{sam}, where $\bigtriangleup $ represents their performance difference with RITM. 
We provide the effect bar graph in \textbf{Figure \ref{fig:ex_3_1}}. 
The results show that CPC-SAM greatly improves the performance of SAM on all three datasets, e.g., the performance on GTEA has increased by more than 10\%. Meanwhile, we provide the OVMS visualization in \textbf{Figure \ref{fig:ex_3_2}} which demonstrates the effectiveness of CPC-SAM.

\subsection{Ablation Studies}
\label{sec:6.3}
In this subsection, we provide the results and analyses of the ablation studies, with more experiments in \textbf{Appendix \ref{sec_app:experiment}}. 

\noindent\textbf{Ablation of Prompt Tuning}
We evaluate the trade-off performance of four optimization strategies on CelebAMask-HQ: (i) $\mathcal{M}_1$: sequentially optimizes CaPL and SAM following existing setting in \textbf{Section \ref{sec:4}}, with \( N_t = 2 \).  (ii) $\mathcal{M}_2$: similar as $\mathcal{M}_1$, but with \( N_t = 4 \).  (iii) $\mathcal{M}_3$: also with \( N_t = 2 \), but optimizes SAM first, followed by CaPL. (iv) $\mathcal{M}_4$: but optimizes CaPL iteratively over an average of 5 rounds. \textbf{Figure \ref{fig:abla_prompt}} shows that $\mathcal{M}_1$ achieves comparable performance compared with $\mathcal{M}_2$ with the lowest calculation overhead. $\mathcal{M}_1$ is also the setting of the current version.

\noindent\textbf{Ablation of Prompt Embeddings}
To assess the benefits of introducing causality, we analyze the impact of various prompts on CelebAMask-HQ. We compare the fixed prompt, learnable prompt, task prompt, and causal prompt (detailed in \textbf{Appendix \ref{sec_app:experiment}}). The learnable prompt lacks causal constraints and is optimized with SAM via soft prompt-tuning \cite{khattak2023maple}. The task prompt only includes the task causal module. \textbf{Figure \ref{fig:abla_1_toy}} shows that the causal prompt achieves the highest dice score than the other prompts, highlighting the significance of leveraging the causality.

\section{Conclusion}
\label{sec:7}
Through empirical and causal analyses, we identify the cause of generalization issues in OVMS as confounders in the prompts, specifically task-irrelevant factors. To address this, we propose CPC-SAM, which eliminates confounders by extracting causal prompts through bi-level optimization, thereby achieving robust OVMS. It integrates CaPL, a lightweight network to derive causal prompts by identifying causal factors within the prompts and then reweighting them through causal multi-distribution consistency. The network is alternately optimized with SAM using bi-level optimization to ensure robust OVMS. Extensive experiments validate the effectiveness of CPC-SAM.
{
    \small
    \bibliographystyle{ieeenat_fullname}
    \bibliography{main}
}

\clearpage
\setcounter{page}{1}

\maketitlesupplementary

\appendix
\section*{Appendix}
The appendix provides supplementary information and additional details that support the primary discoveries and methodologies proposed in this paper. It is organized into several sections: 
\begin{itemize}
    % \item Appendix \ref{sec_app:pseudocode} provides the pseudo-code of CPC-SAM.
    \item Appendix \ref{sec:app_proof} provides a specific definition of the causal prompt, a proof of the theoretical analysis of the main text.
    % \item Appendix \ref{sec_app:discussion} provides a detailed discussion about OVMS settings, the importance of meta-causal prompts, and the advantages of CPC-SAM.
    \item Appendix \ref{sec_app:dataset} provides details for all datasets used in the experiments.
    \item Appendix \ref{sec_app:baseline} provides details for the baselines used in the experiments.
    \item Appendix \ref{sec_app:implementation}  presents the implementation and architecture of our method, aiding in the faithful reproduction of our work.
    \item Appendix \ref{sec_app:metric}  presents the detailed calculation of the evaluation protocols.
    \item Appendix \ref{sec_app:experiment} provides the details of experiments in the main text, full results, additional experiments, and more analyses.
\end{itemize}
Note that before we illustrate the details and analysis, we provide a brief summary of all the experiments conducted in this paper, as shown in Table \ref{tab:app_experiments_summary}.

\begin{table*}[tb]
    \centering
    \begin{tabular}{p{0.4\textwidth}|p{0.3\textwidth}|p{0.2\textwidth}}
    \toprule
        \textbf{Experiments} & \textbf{Location} & \textbf{Results}\\
    \midrule    
        Motivation Experiments & Section \ref{sec:analysis_empirical} and Appendix \ref{sec_app:ex_motivation_overfitting} & Figure \ref{fig:motivation_ex_generalization}, Figure \ref{fig:motivation_ex_prompt_bias}, and Figure \ref{fig:prompt_bias_entity}\\
    \midrule    
        Extension of Motivation Experiments & Appendix \ref{sec_app:ex_motivation_overfitting} & Figure \ref{fig:ex_motivation_causal}\\
    \midrule    
        Experiments on General Target Segmentation with Standard and Few-shot Setting & Section \ref{sec:6.2} & Table \ref{tab:1_common} and Table \ref{tab:1_few-shot}\\
    \midrule    
        Experiments on Medical Segmentation with Standard Setting and Few-shot Setting & Section \ref{sec:6.2} & Table \ref{tab:2}\\
    \midrule    
        Experiments on Complex Scene and Activity Segmentation & Section \ref{sec:6.2} & Figure \ref{fig:ex_3_1} \\
    \midrule    
        Experiments on Out-Of-Distribution (OOD) & Section \ref{sec:6.2} & Table \ref{tab:2}\\
    \midrule    
        Experiments on More Examples & Appendix \ref{sec_app:ex_more_example} & Table \ref{tab:app_more_example}\\
    \midrule    
        Qualitative Results & Appendix \ref{sec_app:ex_visualization} & Figures \ref{fig:app_vis_1}, \ref{fig:app_vis_2}, \ref{fig:app_vis_3}, \ref{fig:app_vis_4}, \ref{fig:app_vis_5}, \ref{fig:app_vis_6}, \ref{fig:app_vis_7}, \ref{fig:app_vis_8}, and \ref{fig:app_vis_9}\\   
    \midrule    
        The Plug-and-Play Nature of CaPL (Empirical Verification of Causal Analysis) & Appendix \ref{sec_app:causal_prompt_different} & Figure \ref{fig:causal_prompt_different}\\   
    \midrule    
        Ablation Study on Prompt Tuning & Section \ref{sec:6.3} & Figure \ref{fig:abla_prompt}\\
    \midrule    
        Ablation Study on Prompt Embeddings & Section \ref{sec:6.3} and Appendix \ref{sec_app:ex_ablation} & Figure \ref{sec_app:experiment}\\
    \midrule    
        Ablation Study on Parameter Sensitivity & Appendix \ref{sec_app:ex_ablation} & Figure \ref{fig:abla_para_xi} and Figure \ref{fig:abla_para_en}\\
    \midrule    
        Ablation Study on Fine-Tuning Method & Appendix \ref{sec_app:ex_ablation} & Figure \ref{fig:abla_fine-tuning}\\
    \midrule    
        Ablation Study on LoRA Layers & Appendix \ref{sec_app:ex_ablation} & Figure \ref{fig:abla_lora}\\
    \bottomrule
    \end{tabular}
    \caption{Illustration of the experiments conducted in this work. All experimental results are obtained after five rounds of experiments.}
    \label{tab:app_experiments_summary}
\end{table*}

\section{Definitions and Proofs}
\label{sec:app_proof}
In this section, we first analyze and give the definition of causal prompts, including structural and optimality conditions, to help better understand the essence of causal prompts. Secondly, we give the detailed proofs of \textbf{Theorem 1} in the main text.

\subsection{Definition of Causal Prompt}
As analyzed in the main text, the generalization capability of segmentation models heavily relies on the quality of prompts, which guide the model to associate input samples with semantic labels. However, conventional prompts often suffer from prompt bias in OVMS—a phenomenon where irrelevant or spurious factors (e.g., environmental noise or annotation artifacts) are inadvertently encoded into the prompts, leading to erroneous correlations between task-irrelevant factors and labels. To address this, we propose the concept of \emph{causal prompt}, grounded in causal inference theory, which explicitly disentangles task-relevant causal factors from irrelevant biases. By formalizing the data generation process through an SCM, we first define the structural requirements for a prompt to strictly encode causal semantics (\textbf{Definition 1}). Subsequently, we characterize its optimality through a minimal-risk objective (\textbf{Definition 2}). Together, these definitions establish a theoretical foundation for designing prompts that enhance model robustness and generalizability in unseen scenarios.

We first provides the definition of \emph{Causal Prompt} based on the notation illustrated in the main text:
\begin{definition}[Causal Prompt]
    Following the causal theory \cite{pearl2009causality} the input \( X \) is generated by task-relevant factors \( F_r^s \) and task-irrelevant factors \( F_{ir}^s \), i.e., $X \coloneqq g(F_r^s, F_{ir}^s), \quad Y \coloneqq h(F_r^s)$
  where \( F_r^s \perp\!\!\!\perp F_{ir}^s \), and \( h(\cdot) \) is a mapping function. A causal prompt \( \text{P} \) must satisfy:
  \begin{itemize}
      \item Causal Representativeness: the causal factors \( F_r^p \) encoded in \( \text{P} \) are semantically equivalent to the causal factors \( F_r^s \) of the input sample, i.e., exists an invertible mapping \( \psi: F_r^s \to F_r^p \) such that $\psi(F_r^s) = F_r^p, \quad \text{and} \quad h(F_r^s) = h \circ \psi^{-1}(F_r^p)$.
      \item Irrelevance Exclusion: \( \text{P} \) is independent of task-irrelevant factors, i.e., $\text{P} \perp\!\!\!\perp F_{ir}^s \quad \text{and} \quad \text{P} \perp\!\!\!\perp F_{ir}^p$.  
  \end{itemize}
\end{definition}

Based on the above definition, intuitively, an optimal causal prompt should not only encode task-relevant causal factors but also minimize the discrepancy between the model's predictions and the true labels across diverse inputs. This requires the prompt to generalize beyond the training data, particularly in open-vocabulary settings where unseen classes and environments are common. To formalize this intuition, grounded on the causal theory \cite{pearl2009causality}, we introduce the optimality condition that defines the causal prompt as the one that minimizes the expected segmentation loss over the data distribution.  
\begin{definition}
    The optimal causal prompt \( \text{P}^* \) is defined as the solution to the following optimization problem:  
\[ \text{P}^* = \arg\min_{\text{P}} \mathbb{E}_{X,Y} \left[ \mathcal{L}\left(Y, f(X; \text{P})\right) \right], \]  
where \( \mathcal{L} \) is the segmentation loss function, and \( f(X; \text{P}) \) is the segmentation model guided by the prompt \( P \).  
\end{definition}
\paragraph{Remark} The optimality condition in \textbf{Definition 2} captures the essence of a causal prompt by focusing on its ability to generalize across diverse and potentially unseen scenarios. By minimizing the expected segmentation loss over the data distribution, the prompt ensures robust performance not only on training data but also in open-vocabulary settings where new classes and environments may arise. While the objective does not explicitly enforce causal alignment, the structural constraints from \textbf{Definition 1} guarantee that the prompt inherently encodes task-relevant causal factors, avoiding spurious correlations with irrelevant features. This dual emphasis on empirical performance and implicit causal alignment makes the causal prompt both theoretically grounded and practically effective, bridging the gap between causal inference and real-world segmentation.

\subsection{Proofs of Theorem 1}

\textbf{\emph{Proof.}} The proof hinges on two pillars: (i) Causal Multi-Distribution Consistency: Random prompts induce diverse confounding perturbations, creating a noise structure that cancels confounder gradients; and (ii) Loss-Driven Invariance: Minimizing $\mathcal{L}(f_\phi^*;X_i,f_\theta(P_i))$ forces the model to rely on stable causal features, as unstable confounders cannot consistently reduce the loss across all $\mathcal{D}_i$. Note that for the convenience of representation, we denote the set of samples $X$ which are consistent in each dataset $\mathcal{D}_i$ as $X_i$.

Based on the causal theory \cite{pearl2009causality,Cheng2017causal}, also analyzed in Subsection \ref{sec:analysis_causal}, the task $\tau$ depends on two components:  (i) invariant causal factors $\mathcal{C}$: Persistent across all perturbed distributions $\{ \mathcal{D}_i \}$, and (ii) environment-specific confounders $\mathcal{E}_i$: Perturbed by random prompt annotations $P_i$.  
In this context, for any input $X_i \in \mathcal{D}_i$, the model observes $X_i = \mathcal{C} \oplus \mathcal{E}_i$, where $\oplus$ denotes the interaction between causal and confounding factors. The optimal model $f_\phi^*$ aims to predict $y_{i,j}$ by disentangling $\mathcal{C}$ from $\mathcal{E}_i$.  
The condition $\mathcal{L}(f_\phi^*; X_i, f_\theta(P_i)) \approx \varepsilon$ implies that the model achieves near-optimal performance across all $\mathcal{D}_i$. Formally, for all $i$:  
\begin{equation}\label{eq:wjy}
\mathbb{E}_{X_i \sim \mathcal{D}_i} \left[ \mathcal{L}(f_\phi^*; X_i, f_\theta(P_i)) \right] \leq \varepsilon.
\end{equation}  
Since $\mathcal{D}_i$ shares the same $X$ but differs in $P_i$, the loss consistency across $\{ \mathcal{D}_i \}$ enforces:  
\begin{equation}
f_\phi^*(X, f_\theta(P_i)) \approx f_\phi^*(X, f_\theta(P_j)), \quad \forall i,j \leq N_t.
\end{equation}  
This alignment requires the model’s predictions to be invariant to prompt-induced perturbations $P_i \rightarrow P_j$, which only affect $\mathcal{E}_i$.  
Next, we provides the gradient analysis of causal and confounding paths. Specifically, let $\theta$ parameterize the prompt optimizer $f_\theta$. The total gradient of $\mathcal{L}$ with respect to $\theta$ is:  
\begin{equation}
\nabla_\theta \mathcal{L} = \sum_{i=1}^{N_t} \nabla_\theta \mathcal{L}(f_\phi^*; X_i, f_\theta(P_i)).
\end{equation}  
Decompose the gradient into contributions from causal and confounding factors:  
\begin{equation}
\nabla_\theta \mathcal{L} = \sum_{i=1}^{N_t} \nabla_\theta \mathcal{L}_{\mathcal{C}} + \sum_{i=1}^{N_t} \nabla_\theta \mathcal{L}_{\mathcal{E}i}.
\end{equation}  
For causal paths, driven by invariant factors $\mathcal{C}$, these gradients reinforce across all $\mathcal{D}_i$ because $\mathcal{C}$ is consistent.  
On the other hand, for confounder gradients ($\nabla_\theta \mathcal{L}_{\mathcal{E}_i}$), driven by perturbed $\mathcal{E}_i$, these gradients are inconsistent across $i$ due to random prompt perturbations (where the property also mentioned in \cite{mohri2018foundations}).  
Under the consistency constraint (Eq.\ref{eq:wjy}), the model must minimize the variance of predictions across $\{ \mathcal{D}_i \}$. This forces $\sum_{i=1}^{N_t} \nabla_\theta \mathcal{L}_{\mathcal{E}_i} \approx 0$.
If confounder gradients do not cancel, predictions would diverge across $\mathcal{D}_i$, violating $\mathcal{L}(f_\phi^*; X_i, f_\theta(P_i)) \approx \varepsilon$. Random perturbations ensure $\mathcal{E}_i$ and $\mathcal{E}_j$ ($i \neq j$) are uncorrelated, making their gradients noisy and canceling in expectation.  
Thus, the optimization of $\mathcal{L}$ suppresses confounder-related updates, leaving only causal gradients to update $\theta$.  
As $\theta$ converges, the prompt optimizer $f_\theta$ learns to map perturbed prompts $P_i$ to a subspace that filters out $\mathcal{E}_i$ and retains $\mathcal{C}$. Formally:  
\begin{equation}
f_\theta(P_i) = \mathcal{C} + \delta_i, \quad \text{where } \delta_i \rightarrow 0.
\end{equation}  
The residual $\delta_i$ vanishes because the loss $\varepsilon$ bounds the impact of $\mathcal{E}_i$. Consequently, $f_\theta(P_i)$ becomes invariant to prompt perturbations, encoding only causal factors.  

Obtaining the above results, we now turn to analyze the generalization issues in OVMS.
For any new prompt $P'$ generated under the same perturbation regime, the causal invariance ensures $f_\theta(P') \approx \mathcal{C}$.  
since $\mathcal{C}$ is the only stable component across all prompts. This guarantees that $f_\theta(P_i)$ generalizes as causal prompts.  

Thus, by enforcing prediction consistency across randomly perturbed prompts, the prompt optimizer $f_\theta$ is compelled to discard environment-specific confounders $\mathcal{E}_i$ and retain invariant causal factors $\mathcal{C}$. The bounded loss $\varepsilon$ ensures that $f_\theta(P_i)$ converges to a representation dominated by $\mathcal{C}$, thereby constituting causal prompts. This aligns with the causal invariance principle and achieves confounder elimination without external data.

\section{Details of Benchmark Datasets}
\label{sec_app:dataset}

\begin{figure*}
\begin{center}
\centerline{\includegraphics[width=\textwidth]{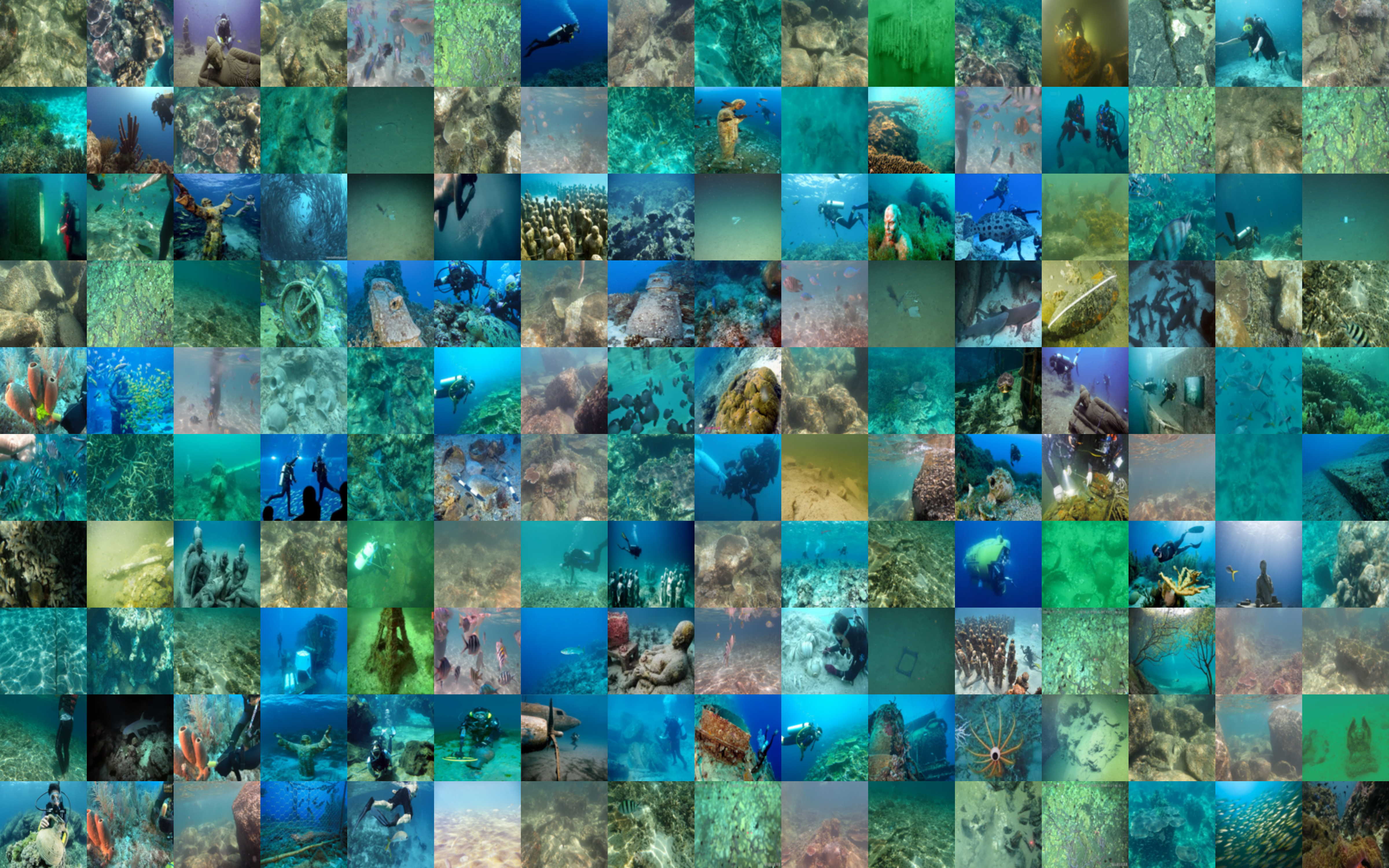}}
\caption{Deep-sea images added to OceanCOCO. These images, taken in challenging environments like the deep sea, inherently suffer from uneven lighting and imbalanced foreground-background proposals, adding complexity to OVMS tasks. Importantly, these images have been pre-processed and included as unlabelled data for OVMS.}
\label{fig_app:oceancoco}
\end{center}
\end{figure*}

Here, we provide a detailed illustration of each benchmark dataset used in the empirical evaluation: 
\begin{itemize}
    \item \textbf{CelebAMask-HQ}\footnote{\url{https://mmlab.ie.cuhk.edu.hk/projects/CelebA/CelebAMask_HQ.html}} \cite{lee2020maskgan} contains 30,000 high-resolution face images with detailed segmentation masks for facial attributes such as eyes, nose, and mouth. We use the last 2,000 examples as the test set.
    \item \textbf{intelecaiCAR}\footnote{\url{https://www.kaggle.com/datasets/intelecai/car-segmentation}} includes 4,073 car images with corresponding segmentation masks. The dataset contains a diverse range of car models and environmental contexts. We split the last 100 examples as the test set.

    \item \textbf{TikTok dances}\footnote{\url{https://www.kaggle.com/datasets/tapakah68/segmentation-full-body-tiktok-dancing-dataset}} comprises 2,615 images of dancing individuals extracted from TikTok videos, with full-body segmentation. All the people were selected in Photoshop. We split the last 2000 examples as the test set.

    \item \textbf{BBBC038v1}\footnote{\url{https://bbbc.broadinstitute.org/BBBC038}} \cite{caicedo2019nucleus} consists of around 200,000 cell images annotated with nuclei segmentation masks. We split the last 5000 examples as the test set.

    \item \textbf{Kvasir-SEG}\footnote{\url{https://datasets.simula.no/kvasir-seg/}} \cite{jha2020kvasir} provides 1,000 endoscopy images annotated with gastrointestinal polyp segmentation masks. We split the last 500 examples as the test set.

    \item \textbf{JSRT}\footnote{\url{https://www.kaggle.com/datasets/raddar/nodules-in-chest-xrays-jsrt}} \cite{shiraishi2000development} contains 247 chest X-ray images annotated with lung nodule segmentation masks. We split the last 147 examples as the test set.

    \item \textbf{GTEA}\footnote{\url{https://cbs.ic.gatech.edu/fpv/}} \cite{fathi2011learning} contains 3,500 frames (7 video sequences) capturing daily activities from a first-person perspective. We follow the original data split for training and evaluation.

    \item \textbf{TrashCan}\footnote{\url{https://conservancy.umn.edu/items/6dd6a960-c44a-4510-a679-efb8c82ebfb7}} \cite{hong2020trashcan} contains 1,484 underwater images labeled with six types of trash, e.g., bottles, cans, etc. We split the last 484 examples as the test set.

    \item \textbf{OceanCOCO} is a self-made dataset that includes thousands of scene images with self-collected deep-sea scenes, covering various underwater objects and activities. On the basis of the most commonly used benchmark, COCO \cite{lin2014microsoft,deng2024coconut}, we add the newly collected deep-sea samples as the target class, including 38 real-world deep-sea species to simulate practical applications. The base and target classes are partitioned into 13:6 to assess generalization. These data are unlabeled, and their raw images come from the ocean benchmark dataset UIEB \cite{li2019underwater} and actual radar images and have been cleaned.  Figure \ref{fig_app:oceancoco} shows part of the deep-sea images. We split the last 20 examples for each class as the test set.
\end{itemize}
Note that each dataset is split into annotated base classes and unannotated target classes in a $2:1$ ratio. What we want to emphasize here is that, unlike previous experiments on conventional segmentation tasks (posted in the original papers), we follow the OVMS setting, which focuses on the performance of the model on unknown new classes and multiple entities. Even so, our performance on OVMS even exceeds that of these baselines on high-quality supervised data.

\section{Details of Baselines}
\label{sec_app:baseline}
We briefly introduce the baselines used in the experiments.
\begin{itemize}
    \item DeepLabV3 \cite{chen2017rethinking}: A widely used supervised method for semantic segmentation, which introduces atrous spatial pyramid pooling (ASPP) to capture multi-scale context.
    \item SwinUnet \cite{cao2022swin}: An innovative U-Net architecture that leverages the Swin Transformer to enhance segmentation performance by utilizing a hierarchical structure with shifted windows for efficient context learning.
    \item RITM \cite{sofiiuk2022reviving}: proposes a feedforward model for click-based interactive segmentation that employs the segmentation masks from previous steps.
    \item ViTDet-H \cite{li2022exploring,zhang2024exploringroletokentransformerbased}: A vision transformer-based detection model that explores transformer architectures for segmentation, leveraging self-attention mechanisms for precise spatial relationship modeling.
    \item HSNet \cite{zhang2022hsnet}: uses hierarchical features to generate prototypes for target classes, and uses cross-semantic attention to bridge the gap between low-level and high-level features.
    \item SSP \cite{fan2022self}: uses query prototypes from limited data and high-confidence query predictions to match query features, enabling the model to generalize better with limited samples through effective feature reuse.
    \item Vanilla SAM \cite{sam}: The foundational model that focuses on generalized, promptable segmentation, capable of handling diverse inputs and domains.
    \item Med-SA \cite{wu2023medical}: A medical segmentation adaptation of SAM, proposes Space-Depth Transpose to adapt 2D SAM to 3D medical images and Hyper-Prompting Adapter to achieve prompt-conditioned adaptation..
    \item SAMed \cite{tancik2020fourier}: Incorporates Fourier analysis to enhance the SAM model for medical image segmentation, introducing frequency-domain information to improve robustness and accuracy in segmentation tasks.
    \item BLO-SAM \cite{zhangblo}: An extension of SAM, optimizing both the segmentation network and task-specific objectives to address the limitations of human annotation.
    \item OVSAM \cite{yuan2024open}: A SAM-based method targeting OVMS, incorporating CLIP to enhance SAM's ability.
    \item ZS3Net \cite{bucher2019zero}: A zero-shot segmentation model designed for OVMS, utilizing textual embeddings from pre-trained language models to generalize to unseen categories.
    \item LSeg \cite{li2022language}: A language-supervised segmentation model that aligns visual and linguistic features, leveraging VLMs to provide zero-shot segmentation capabilities. 
    \item ZegFormer \cite{ding2022decoupling}: A transformer-based model for OVMS, introducing a decoupling mechanism that separates class-agnostic segmentation from category-specific predictions to improve performance in unseen categories.
    \item OpenSeg \cite{ghiasi2022scaling}: A model that focuses on scaling OVMS to large datasets by integrating multimodal models, combining image and text features to improve the generalization.
    \item OVSeg \cite{liang2023open}: An OVMS model that enhances scalability and generalization by incorporating a more advanced vision-language alignment strategy.
\end{itemize}
These innovations contribute to the diversity of segmentation techniques, ranging from supervised to advanced open-vocabulary and few-shot segmentation models.

\section{Details of Implementation}
\label{sec_app:implementation}

Regarding the network architecture, CPC-SAM comprises two components: CaPL and SAM, which are optimized alternately during training. SAM utilizes the ViT-B model as its image encoder, while the CaPL employs a simple 2-layer MLP to calibrate prompts.
For SAM, we follow the original SAM setup. For the image encoder, we use the Vision Transformer (ViT) \cite{dosovitskiy2020image} pre-trained by MAE \cite{he2022masked}, specifically ViT-B, which features $14 \times 14$ window attention and four equidistant global attention blocks for adjustment. Note that SAM has three model versions with different types of image encoders, ranging from ViT-B to ViT-L and ViT-H. We choose ViT-B, representing the smallest encoder model, to improve efficiency. Unlike the original SAM, for high-resolution datasets such as OceanCOCO, we rescale the images twice, then pad the shorter edges, followed by applying $1 \times 1$ and $3 \times 3$ convolutions. For CaPL, we introduce a lightweight two-layer convolutional structure to iteratively extract causal variables based on the original convolutional network structure. This network is optimized alternately with the segmentation model; simply put, the causal learner is frozen when the SAM-based framework is optimized. The sparse prompts obtained are mapped into 256-dimensional vector embeddings, where each point is encoded by its position and the sum of one of two learned embeddings (for foreground or background prompts) and the causal weighting vector. Finally, we use the transformer MLP blocks for the mask decoder but use LoRA layers for prompt-tuning. More details about the PyTorch version architecture and experimental checkpoints will be released in the final version.

For implementation, we use eight A100 GPUs with 80G memory for distributed training. Each GPU processed two images per mini-batch. For supervised and few-shot methods, we use the parameter settings provided in the original papers. For SAM-based and OVMS-specific models, we use Adapter \cite{houlsby2019parameter} and LoRA \cite{hu2021lora} for SAM fine-tuning. Similarly, the LoRA layers and unfrozen components are optimized using the Adam optimizer, with an initial learning rate of $5e-3$, betas set to $(0.9, 0.999)$, and a weight decay of $0.1$. Specifically, all trainable parameters are updated on a single training set based on the configurations published in the original papers. 
During inference, since the model requires user-generated prompts, it is impractical to use this approach when testing on many images. Hence, following the evaluation protocol of SAM and Med-SA, it is unrealistic to derive prompts from ground-truth masks in advance. Unlike previous works, we do not require manual priors during the testing phase. Instead, we use a 10\% subset of the data to generate prompts.

\section{Details of Evaluation Protocols}
\label{sec_app:metric}

In this section, we introduce the evaluation metric used to assess the performance of the model: Dice score. Meanwhile, we also consider the IoU for the evaluations on complex scene and activity segmentation following \cite{sam}.

\textbf{Intersection over Union (IoU)}
Intersection over Union (IoU) is used to measure the segmentation accuracy by comparing the predicted mask with the ground truth mask. It is defined as the ratio of the intersection of the predicted and ground truth masks to their union:
\begin{equation}
{Score}_{\text{IoU}} = \frac{|A \cap B|}{|A \cup B|}
\end{equation}
where $A$ and $B$ are the predicted and ground truth masks.

\textbf{Dice Score}
The Dice score is another metric for evaluating the overlap between the predicted and ground truth masks, particularly emphasizing the similarity between them. The Dice score is calculated as:
\begin{equation}
{Score}_{\text{Dice}} = \frac{2|A \cap B|}{|A| + |B|}
\end{equation}
where $A$ and $B$ represent the predicted and ground truth masks, respectively. It can be regarded as the consideration of both accuracy and IoU score, which is also the selection in our work. The Dice score ranges from 0 to 1.

\section{Additional Results}
\label{sec_app:experiment}

This section presents additional experimental details, quantitative analyses, full results, and omitted experiments due to space constraints. Specifically, Appendix \ref{sec_app:ex_motivation_overfitting} provides full details and results of the motivation experiments. Appendix \ref{sec_app:ex_ablation} presents additional ablation studies, including parameter sensitivity, fine-tuning methods, and the rank of the LoRA layers.
Appendix \ref{sec_app:ex_more_example} examines method performance on larger sample sizes, extending the standard and few-shot settings discussed in the main text. 
Finally, we visually validate CPC-SAM's effectiveness and robustness in OVMS tasks through segmentation mask visualizations across nine datasets in Appendix \ref{sec_app:ex_visualization}.

\subsection{Details of Motivating Experiments}
\label{sec_app:ex_motivation_overfitting}

In Section \ref{sec:analysis_empirical} of the main text, we construct three sets of experiments for the existence and exploration of generalization issues in OVMS. Here, we further supplement some details of these experiments, e.g., the insights behind each experiment, what we can conclude from the results, and the results after introducing causal prompts.

\paragraph{Experiments for Existence of Generalization Issue} 
In OVMS, models must generalize to unseen classes under significant distribution shifts between training (base classes) and testing (target classes). While models like SAM exhibit strong adaptability in standard segmentation tasks, their robustness in OVMS scenarios—particularly for rare or novel categories—remains uncertain. To address this, we construct the OceanCOCO dataset, a real-world benchmark comprising deep-sea samples with rare species classes (\textbf{Figure \ref{fig_app:oceancoco}}). These classes simulate practical applications where segmentation models encounter entities rarely annotated in existing datasets (e.g., deep-sea exploration or ecological monitoring).  
OceanCOCO explicitly splits classes into a 13:6 ratio of base (seen) to new (unseen) classes, intentionally inducing a distribution shift that mirrors real-world imbalance. 
The choice of rare deep-sea species reflects practical scenarios in deep-sea exploration where segmentation models must handle underrepresented or novel entities. Such environments inherently involve limited prior annotations and high-class imbalance, making them ideal for stress-testing generalization. 
By fine-tuning models on base classes and evaluating performance on both splits, we directly probe their ability to generalize in OVMS. The observed trend (Figure \ref{fig:motivation_ex_generalization})—improving performance on base classes but declining accuracy on new classes after initial training—reveals a critical insight: standard training protocols risk overfitting to seen categories, even for highly adaptable models like SAM. This decline highlights the challenge of OOD generalization issues.

\paragraph{Experiments for Exploration of Generalization Issue}
This experiment aims to isolate and quantify the impact of prompt quality for SAM on the OOD generalization issues in OVMS tasks. 
In OVMS, the model may encounter many classes absent during training, causing a substantial distribution shift between training and testing. This shift makes it challenging for a model trained on base classes to generalize to unseen target classes, reflecting OOD generalization issues. Thus, we think the aforementioned generalization issues are mainly from the OOD problem in SAM.
To evaluate what causes this issue, we begin by analyzing what impacts the performance of SAM. As mentioned in the main text.
Generally, the performance of SAM depends on two factors: the pre-trained network and the prompt. According to existing study about scaling law and VLMs \cite{zhou2023limaalignment,ye2025limoreasoning}, the network of SAM has pre-trained on 11 million unlabeled images, which has learned cross-domain feature representations. Thus, the pre-trained network is considered to be able to generalize to new classes. This implies that the pre-trained network is reasonably effective in handling OOD challenges.  
In contrast, the prompt may play a more crucial role in the OOD generalization issues. The SAM-based methods mainly rely on two types of prompts:  manually interactive prompts and generated prompts. However, in OVMS, the manually interactive prompts (e.g., marking points and bounding boxes) rely on expert knowledge and may struggle to capture fine-grained attributes of unseen classes; generative prompts depend on the quality of generative models, which can result in a semantic mismatch between the prompt and the image. 

To explore this point, we test two critical hypotheses. First, whether refining prompts improves OOD generalization. By comparing fixed generative prompts (prone to semantic mismatch due to training on base classes) with expert-refined prompts (manually adjusted to exclude biases), we isolate the role of prompt precision. The 32.4\% performance gap between these two prompt sets (Figure \ref{fig:motivation_ex_prompt_bias}) directly links prompt quality to segmentation accuracy. This suggests that even minor semantic misalignments in prompts disproportionately degrade performance on rare classes, highlighting prompt design as a key bottleneck. Second, we investigate whether prompt bias affects entities differently within the same sample. By visualizing segmentation results for multi-entity samples, we observe that the same prompt may accurately segment common entities while failing on rare ones (Figure \ref{fig:prompt_bias_entity}). This heterogeneity implies that prompt bias is context-dependent: prompts may overemphasize features shared by frequent classes (e.g., texture or shape priors) while neglecting discriminative attributes of rare entities. Such biases amplify errors in OVMS, where samples often contain diverse and imbalanced entities. This is also the reason why we design an additional entity causal module in CPC-SAM.

The experiment underscores that SAM’s generalization failure in OVMS is not a limitation of its visual backbone but a consequence of suboptimal prompts. Expert-refined prompts act as proxies for ideal causal prompts, manually filtering out irrelevant factors and aligning with causal semantics. The performance gap and entity-level variability jointly motivate the need for automated methods to generate prompts that rigorously encode task-relevant factors while suppressing biases. This insight lays the groundwork for designing causal prompts, which aim to replicate the precision of expert knowledge without human intervention, ensuring robustness across both common and rare classes.

\begin{figure}
\begin{center}
\centerline{\includegraphics[width=\linewidth]{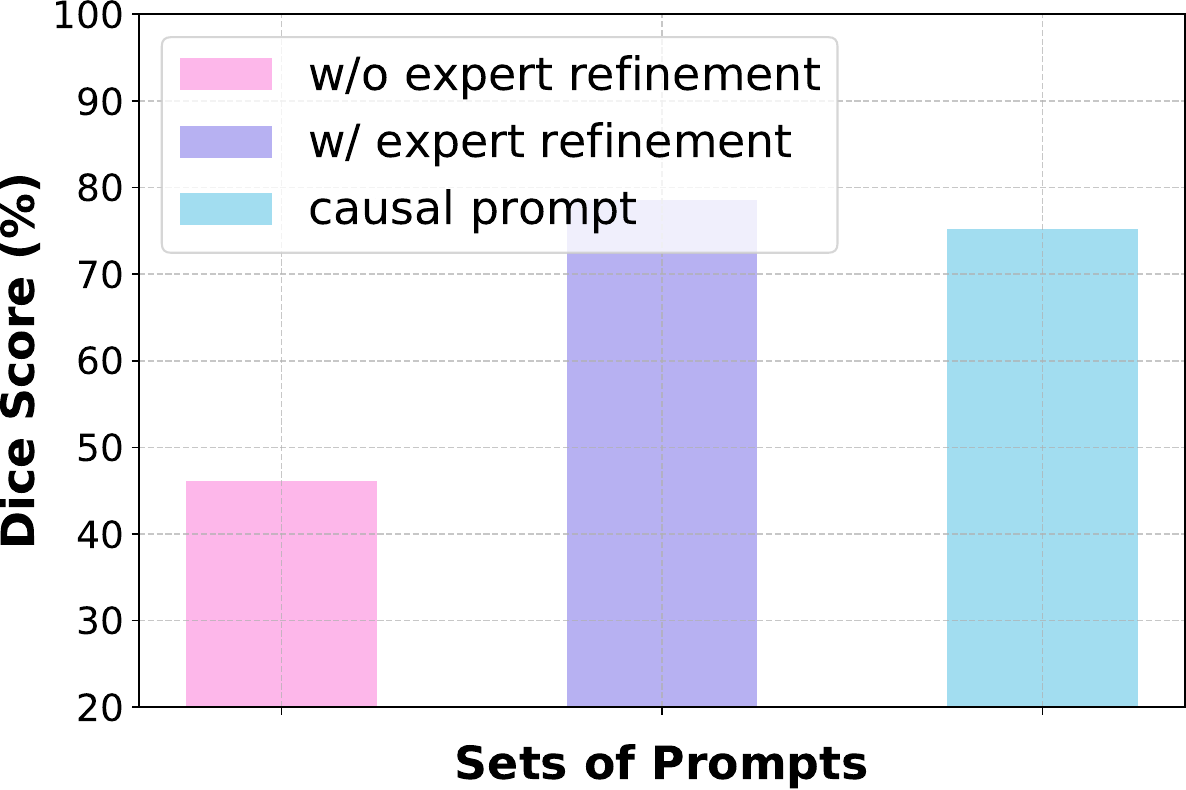}}
\caption{Comparison of three types of prompts.}
\label{fig:ex_motivation_causal}
\end{center}
\end{figure}

\paragraph{Results After Introducing Causal Prompt}
To evaluate the effect of introducing the causal prompt, we conduct an assessment using the same experimental settings of the above motivating experiments. Specifically, we used the same test set as in the previous experiment about the exploration of generalization issues, consisting of 60 randomly sampled instances from OceanCOCO, covering 12 classes.
Next, building on the two previously used prompt types—(i) prompts generated by a fixed pre-trained prompt generator \cite{yuan2024open} and (ii) prompts refined with expert knowledge—we introduced the causal prompt. The causal prompts are generated by feeding each sample into the trained CaPL for reweighting. Finally, we evaluate the segmentation performance of SAM under these three prompt conditions. The results are shown in \textbf{Figure \ref{fig:ex_motivation_causal}}. From the results, we can observe that the causal prompts achieve performance comparable to expert-refined prompts and significantly outperform those generated by a pre-trained model. This demonstrates that the causal prompt can effectively reduce OOD generalization issues in OVMS. It also demonstrates the importance of causality and the advantages of our analyses.

\begin{figure}
\begin{center}
\centerline{\includegraphics[width=\linewidth]{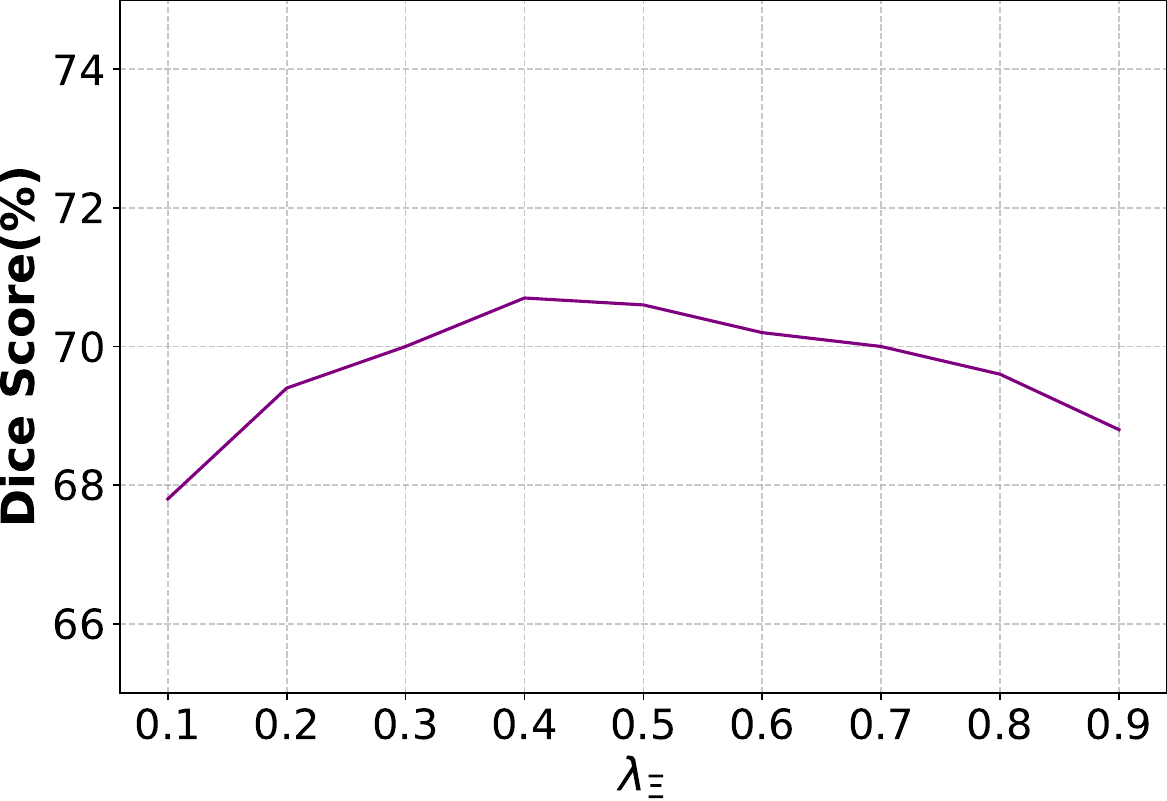}}
\caption{Ablation study of the trade-off hyper-parameter $\lambda_{\Xi}$}
\label{fig:abla_para_xi}
\end{center}
\end{figure}

\begin{figure}
\begin{center}
\centerline{\includegraphics[width=\linewidth]{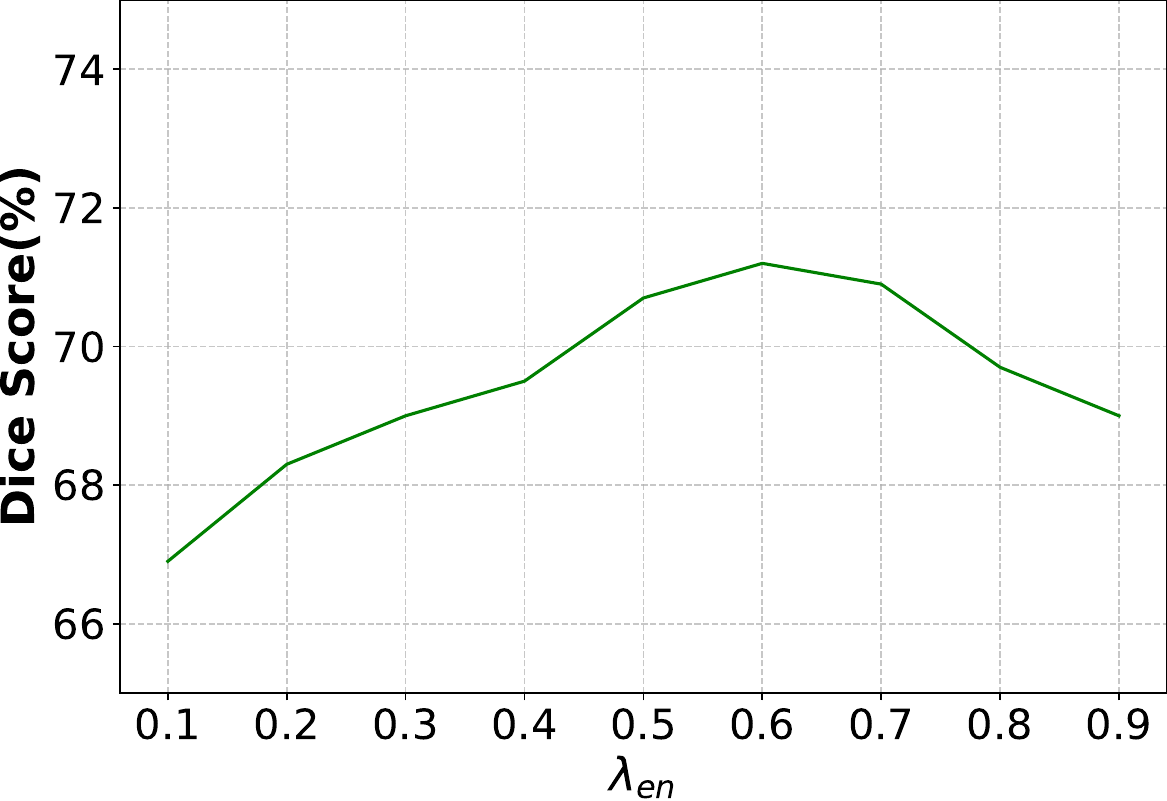}}
\caption{Ablation study of the trade-off hyper-parameter $\lambda_{en}$}
\label{fig:abla_para_en}
\end{center}
\end{figure}

\begin{figure}
\begin{center}
\centerline{\includegraphics[width=\linewidth]{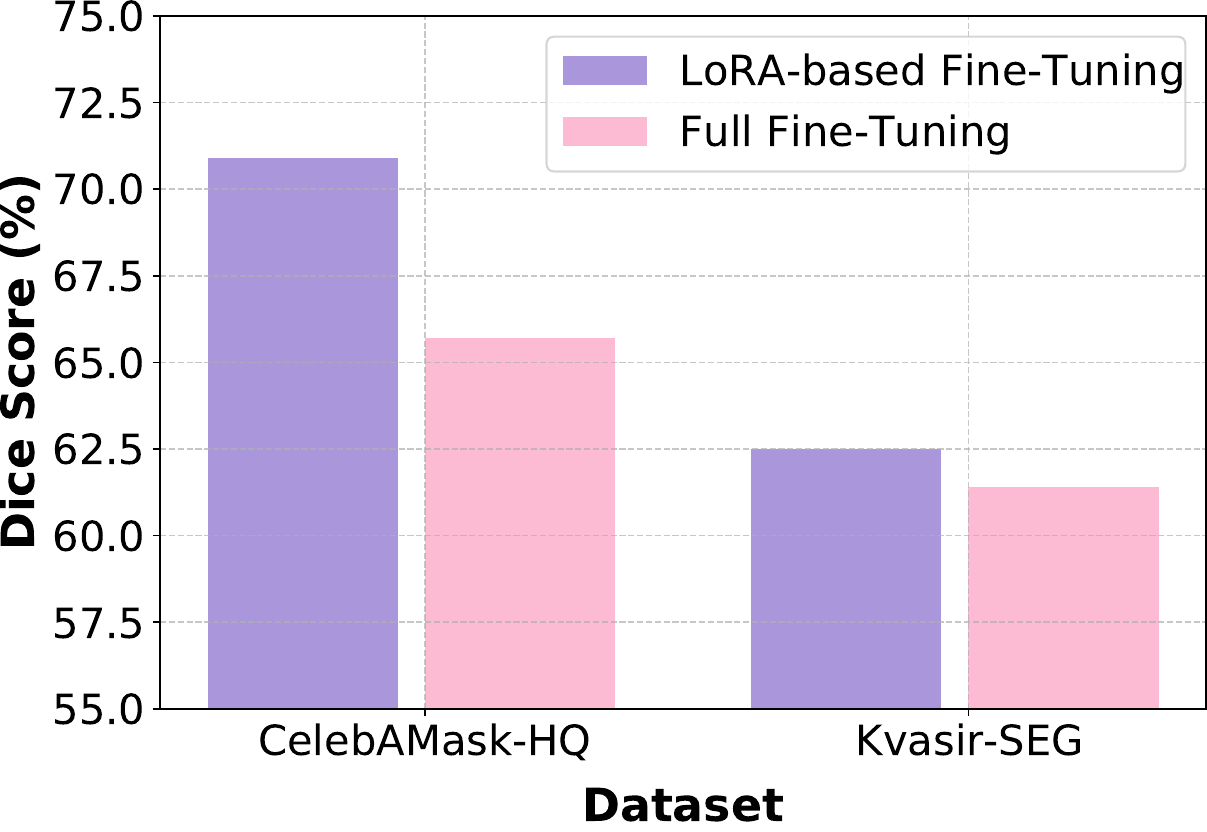}}
\caption{Ablation study of the fine-tuning method within CPC-SAM, i.e., LoRA-based fine-tuning vs. full fine-tuning.}
\label{fig:abla_fine-tuning}
\end{center}
\end{figure}

\begin{figure}
\begin{center}
\centerline{\includegraphics[width=\linewidth]{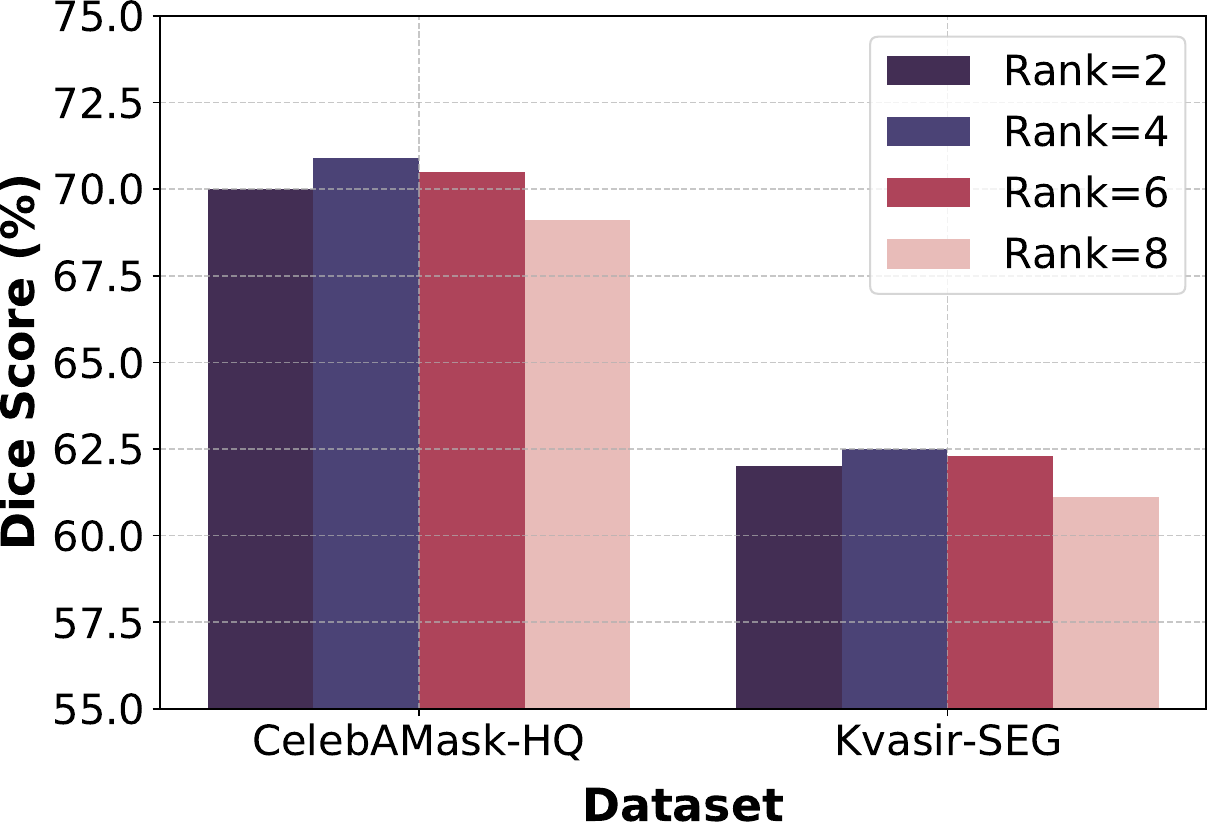}}
\caption{Ablation study of LoRA layers' rank. }
\label{fig:abla_lora}
\end{center}
\end{figure}

\subsection{Other Ablation Studies}
\label{sec_app:ex_ablation}

\paragraph{Parameter Sensitivity Analysis}

We construct two sets of experiments to analyze how different settings of the hyper-parameters $\lambda_{\Xi}$ and $\lambda_{en}$ affect the performance of CPC-SAM. Specifically, we evaluate the performance of CPC-SAM with different hyperparameters following the same settings in Section \ref{sec:6.2}. The results for the two trade-off hyperparameters are shown in Figure \ref{fig:abla_para_xi} and Figure \ref{fig:abla_para_en}, respectively. The results show that $\lambda_{\Xi}=0.4$, $\lambda_{en}=0.6$ can bring the best performance, which is also our setting.

\paragraph{Ablation Study of Full Fine-tuning}

Recall the framework of the proposed CPC-SAM, the second-level update of CPC-SAM employs LoRA layers while keeping other parameters fixed, a strategy known as LoRA-based fine-tuning. This approach aims to reduce resource consumption and mitigate catastrophic forgetting during few-shot fine-tuning. To evaluate the advantages of this approach, we conduct experiments under identical conditions by replacing the second-level LoRA-assisted update with a full update of all SAM components—referred to as full fine-tuning. Notably, this process retains CaPL while fine-tuning based on causal prompts. We compare the performance of CPC-SAM under LoRA-based fine-tuning and full fine-tuning, using the CelebAMask-HQ and Kvasir-SEG benchmark datasets under the same settings as Section \ref{sec:6.2}. As shown in Figure \ref{fig:abla_fine-tuning}, the model's performance slightly declines with full fine-tuning, indicating that LoRA-based fine-tuning provides performance benefits. Moreover, Table \ref{tab:1_few-shot} and Table \ref{tab:2} show that even with full fine-tuning, CPC-SAM surpasses Vanilla SAM by over 30\%, further underscoring the impact of CaPL.

\paragraph{Rank of LoRA Layers}

The rank in LoRA layers determines the capacity of low-rank adaptation during fine-tuning. Higher ranks capture complex patterns, enhancing fine-tuning on tasks requiring significant adaptation, while lower ranks reduce computational costs and suit resource-limited settings. Thus, for CPC-SAM, choosing an appropriate rank enables efficient, effective fine-tuning, especially with limited data or resources. We conduct an ablation study to determine the rank within LoRA layers.
Specifically, we choose the CelebAMask-HQ and Kvasir-SEG benchmark datasets for analysis. We set the rank in the range of \([2,8]\) and evaluate the model's performance with different rank values in LoRA layers. The results are shown in Figure \ref{fig:abla_lora}. It indicates that the optimal performance is achieved with \(Rank=4\) on both datasets. Moreover, although there are performance differences when setting different ranks, the variations are not significant, suggesting that CPC-SAM is not affected by model complexity and avoids overfitting that may result from a high rank.

\paragraph{Details of Prompt Embedding Ablation Study}

To evaluate the impact of introducing causal prompts, we conducted an experiment on prompt embeddings in Section \ref{sec:6.3} of the main text. Specifically, we compared four types of prompt embeddings: fixed, learnable, task, and causal prompts. Below, we provide a brief overview of these prompt embeddings and their differences.
The fixed prompt originates from SAM \cite{sam} and remains unchanged throughout the task. It includes points, boxes, and masks, which integrate different levels of user input or prior information, that is, reconstructed from segmentation masks and text labels in the dataset. Points designate regions to include or exclude, boxes define approximate object boundaries, and masks provide pixel-level guidance.
The learnable prompt, inspired by SAMed \cite{tancik2020fourier}, is optimized jointly with other model parameters during training. Initialized as learnable vectors, the prompt is updated through backpropagation along with the loss function, allowing it to adapt and effectively guide the model to task-relevant features. This learnable prompt serves as auxiliary contextual information, coupling tightly with input features.
The task and causal prompts are both derived from the CaPL of CPC-SAM during the first-level optimization. The task prompt relies solely on the task causal module, focusing on extracting task-relevant factors based on causal invariance. In contrast, the causal prompt is derived from both the task causal and entity causal modules, accounting for relationships between entities, which is particularly beneficial for OVMS.

\begin{table}
\centering
\caption{Performance comparison (average Dice score (\%) with standard deviations) on CelebAMask-HQ and Kvasir-SEG with more training examples. $N$ denotes the number of labeled examples. The best results are highlighted in \textbf{bold}.}
\label{tab:app_more_example}
    \resizebox{1\linewidth}{!}{
    \begin{tabular}{lcccc}
\toprule
    \multirow{2.5}{*}{Method} & \multicolumn{2}{c}{CelebAMask-HQ} & \multicolumn{2}{c}{Kvasir-SEG}  \\
    \cmidrule(lr){2-3} \cmidrule(lr){4-5} 
    & $N_b=128$ & $N_b=512$ & $N_b=128$ & $N_b=512$ \\
\midrule
    Med-SA & 82.8 $\pm$ 1.3 & 85.7 $\pm$ 1.0 & 77.8 $\pm$ 1.3 & 84.0 $\pm$ 0.8 \\
    SAMed & 77.9 $\pm$ 0.9 & 81.7 $\pm$ 1.1 & 72.9 $\pm$ 0.9 & 78.6 $\pm$ 0.9 \\
    BLO-SAM & 86.4 $\pm$ 1.2 & 88.1 $\pm$ 0.8 & 85.9 $\pm$ 0.5 & 88.2 $\pm$ 0.7 \\
    OVSAM & 84.7 $\pm$ 1.2 & 90.2 $\pm$ 1.4 & 81.6 $\pm$ 0.9 & 87.9 $\pm$ 1.0 \\
    CPC-SAM & \textbf{91.6 $\pm$ 1.0} & \textbf{93.8 $\pm$ 1.4} & \textbf{89.4 $\pm$ 1.1} & \textbf{91.0 $\pm$ 1.1} \\
\bottomrule
\end{tabular}}
\end{table}

\subsection{Performance on More Examples}
\label{sec_app:ex_more_example}

Although the experiments in the main text consider both standard and few-shot settings, the sample size in the standard setting is still relatively small compared to the training configurations and batch sizes of other baselines in their original papers \cite{sam,wu2023medical,tancik2020fourier}. Therefore, we conduct experiments to explore how increasing the number of training samples impacts model performance across different tasks. For analysis, we select CelebAMask-HQ and Kvasir-SEG as benchmark datasets from the general target segmentation and medical segmentation domains, respectively. We increase the training sample size to 128 and 512 and conduct experiments with Med-SA, SAMed, BLO-SAM, OVSAM, and CPC-SAM. The results, shown in Table \ref{tab:app_more_example}, indicate that CPC-SAM's performance further improves as the number of training samples increases. Specifically, it achieves the highest Dice scores in all experiments, regardless of whether 128 or 512 labeled samples were used for training. This highlights the importance of sample size in extracting causal factors. Notably, even when Med-SA and SAMed are trained with 512 samples, their performance does not surpass CPC-SAM's performance with just 128 samples, further demonstrating CPC-SAM's causal-based advantage when dealing with data scarcity.

\subsection{The Plug-and-Play Nature of CaPL}
\label{sec_app:causal_prompt_different}
As detailed illustrated in the main text, in the CPC-SAM framework, the lightweight independent component, CaPL, once embedded into the SAM architecture, autonomously achieves causal prompt calibration through a two-stage decoupled optimization. In the first stage, the SAM backbone remains fixed, and CaPL filters out causal features relevant to semantic labels from confounded prompts by leveraging multi-distribution random prompt generation and loss-driven reweighting. In the second stage, with CaPL parameters locked, only the calibrated causal prompts are fed into SAM for fine-tuning to enhance open-vocabulary segmentation performance. This design enables CaPL to dynamically eliminate prompt bias without modifying SAM's original architecture or relying on external knowledge bases. Through modular embedding and phased optimization, it achieves zero-cost adaptation, significantly improving the model's generalization capability for unseen categories. Thus, our CaPL in CPC-SAM has a plug-and-play nature and can be embedded into any SAM-based model.

\begin{figure}
\begin{center}
\centerline{\includegraphics[width=\linewidth]{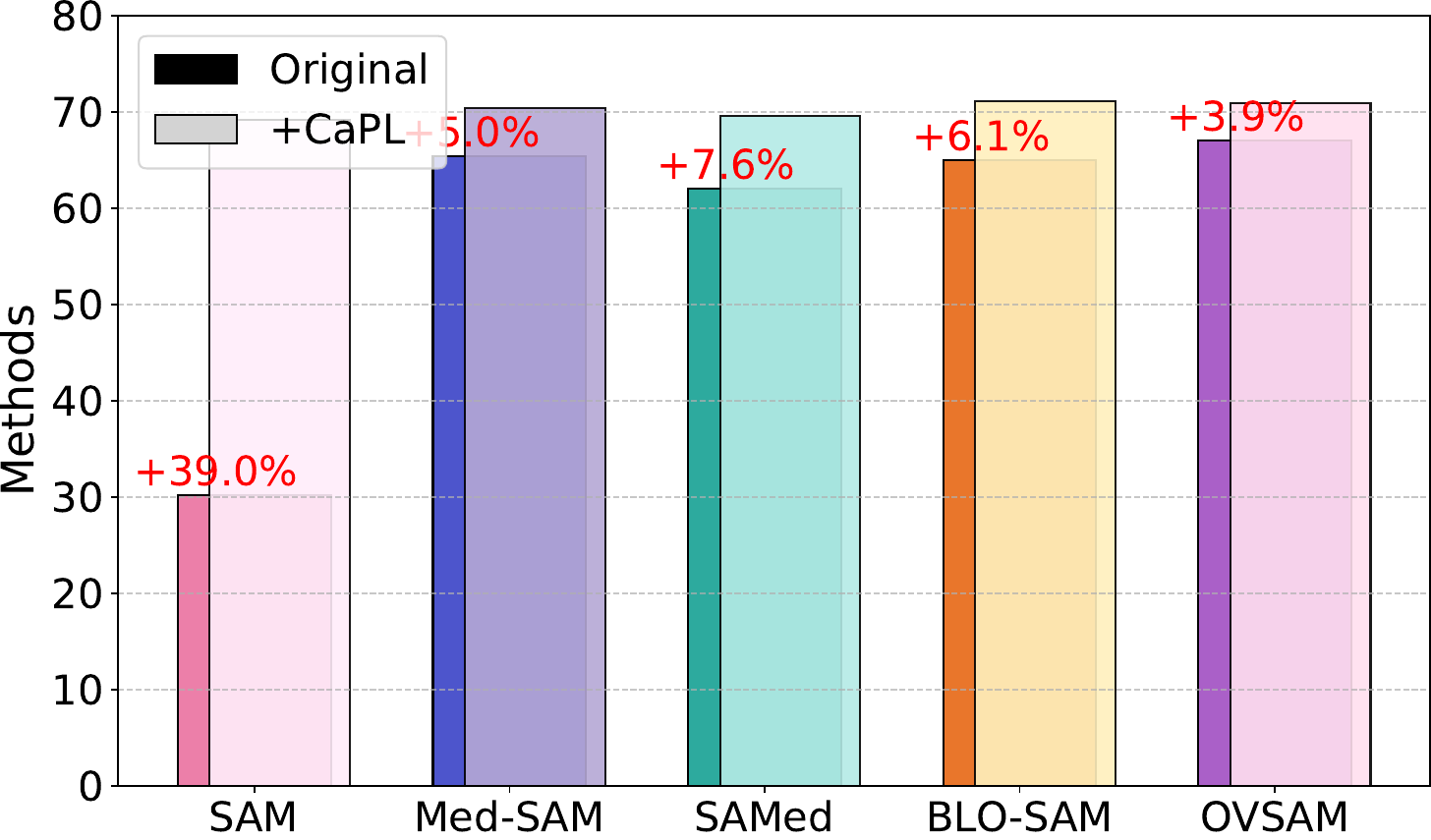}}
\caption{Impact of CaPL on various SAM-based methods. }
\label{fig:causal_prompt_different}
\end{center}
\end{figure}

While the extensive experiments discussed earlier have demonstrated the superiority of CaPL, we conduct further exploration and analysis in this section to verify that the performance improvement indeed stems from extracting causal factors while removing confounders, as well as to validate its plug-and-play nature. Specifically, to evaluate the role of CaPL—or more broadly, this novel causality-based prompt calibration—across various SAM variants, we designed a set of experiments. Adopting the same experimental setup as in Section \ref{sec:6.2}, we assessed the performance changes of existing baselines after integrating CaPL calibration. We used CelebAMask-HQ as the benchmark dataset. Notably, for baseline methods such as SAMed and BLO-SAM, which already include prompt optimization modules, we inserted CaPL after their original prompt optimization steps to perform secondary calibration. If the integration of CaPL improved model performance, this would indicate that the original baselines failed to learn truly causal prompts and were suboptimal. In other words, the significant performance gains of CPC-SAM are indeed attributable to the accurate extraction of causal factors and the elimination of confounding effects. The experimental results, shown in \textbf{Figure \ref{fig:causal_prompt_different}}, reveal that compared to the original baselines, models equipped with CaPL achieved substantial improvements, with an average gain of at least 3.9\%. This validates the importance of causal prompts, the effectiveness of calibration, and the reliability of our causal analysis.

\subsection{Qualitative Results}
\label{sec_app:ex_visualization}

\begin{figure*}
\begin{center}
\centerline{\includegraphics[width=\textwidth]{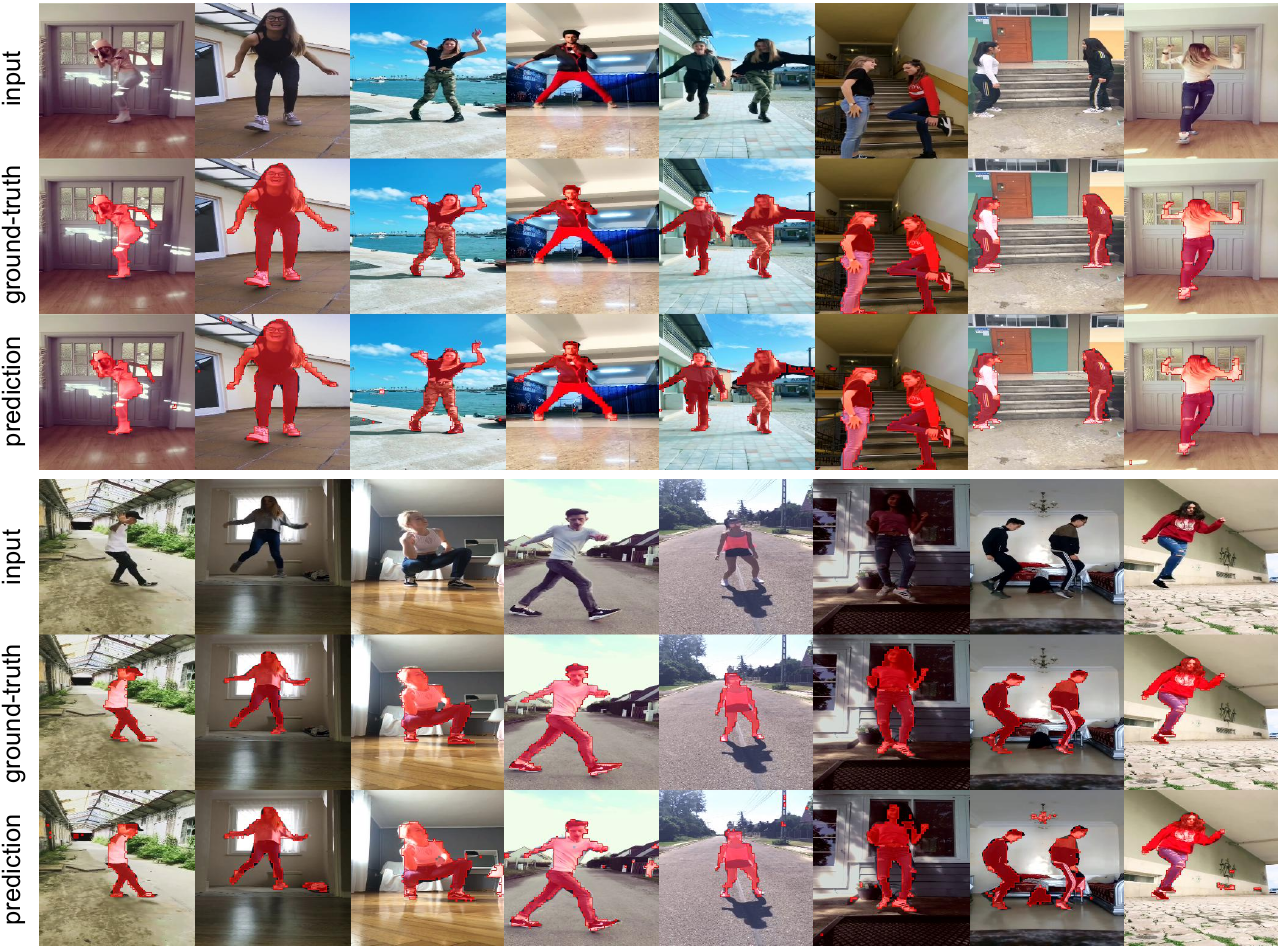}}
\caption{Qualitative results on some randomly sampled test examples from the TikTok dances dataset.}
\label{fig:app_vis_1}
\end{center}
\end{figure*}

\begin{figure*}
\begin{center}
\centerline{\includegraphics[width=\textwidth]{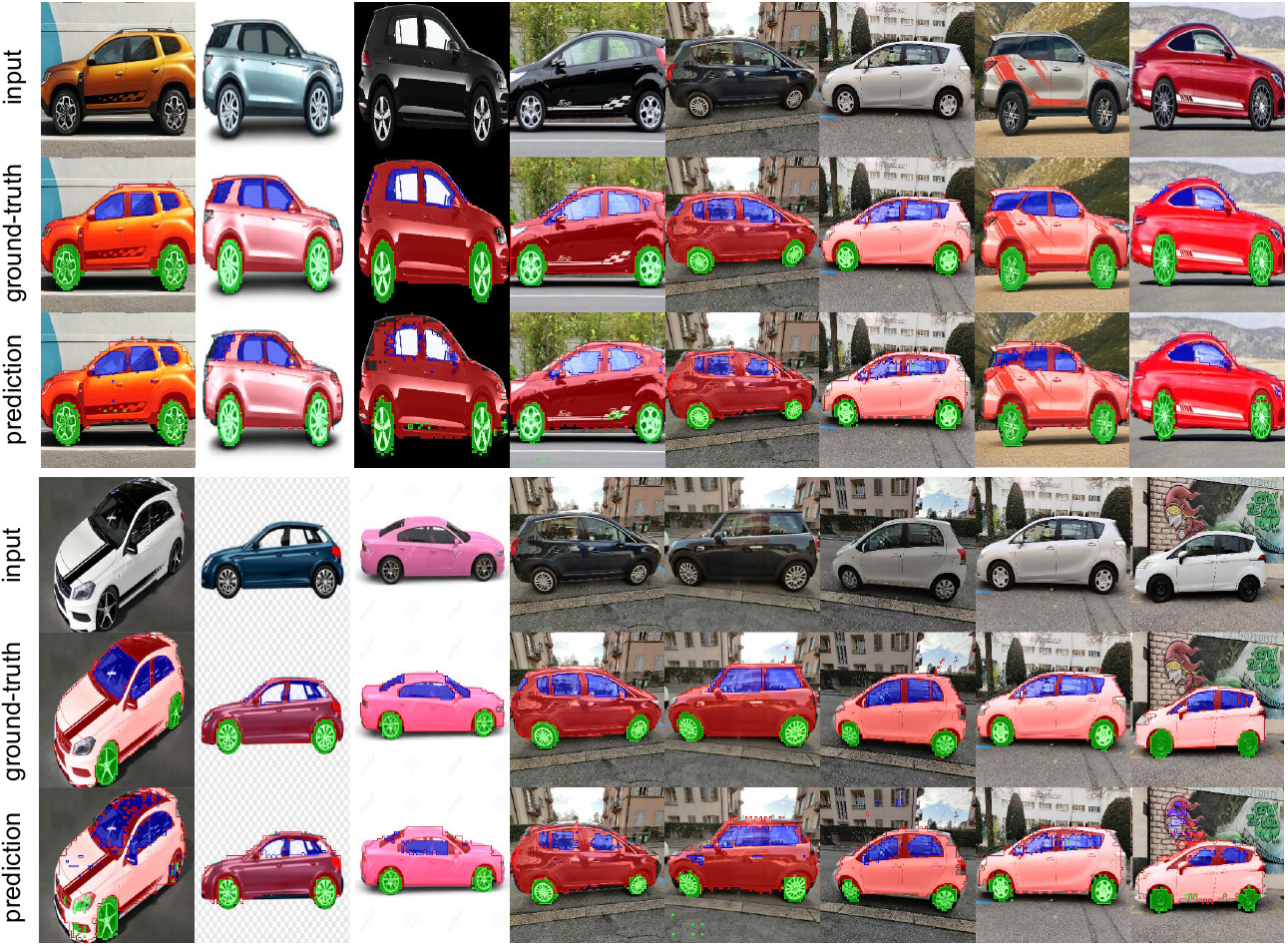}}
\caption{Qualitative results on some randomly sampled test examples from the intelecaiCAR dataset.}
\label{fig:app_vis_2}
\end{center}
\end{figure*}

\begin{figure*}
\begin{center}
\centerline{\includegraphics[width=\textwidth]{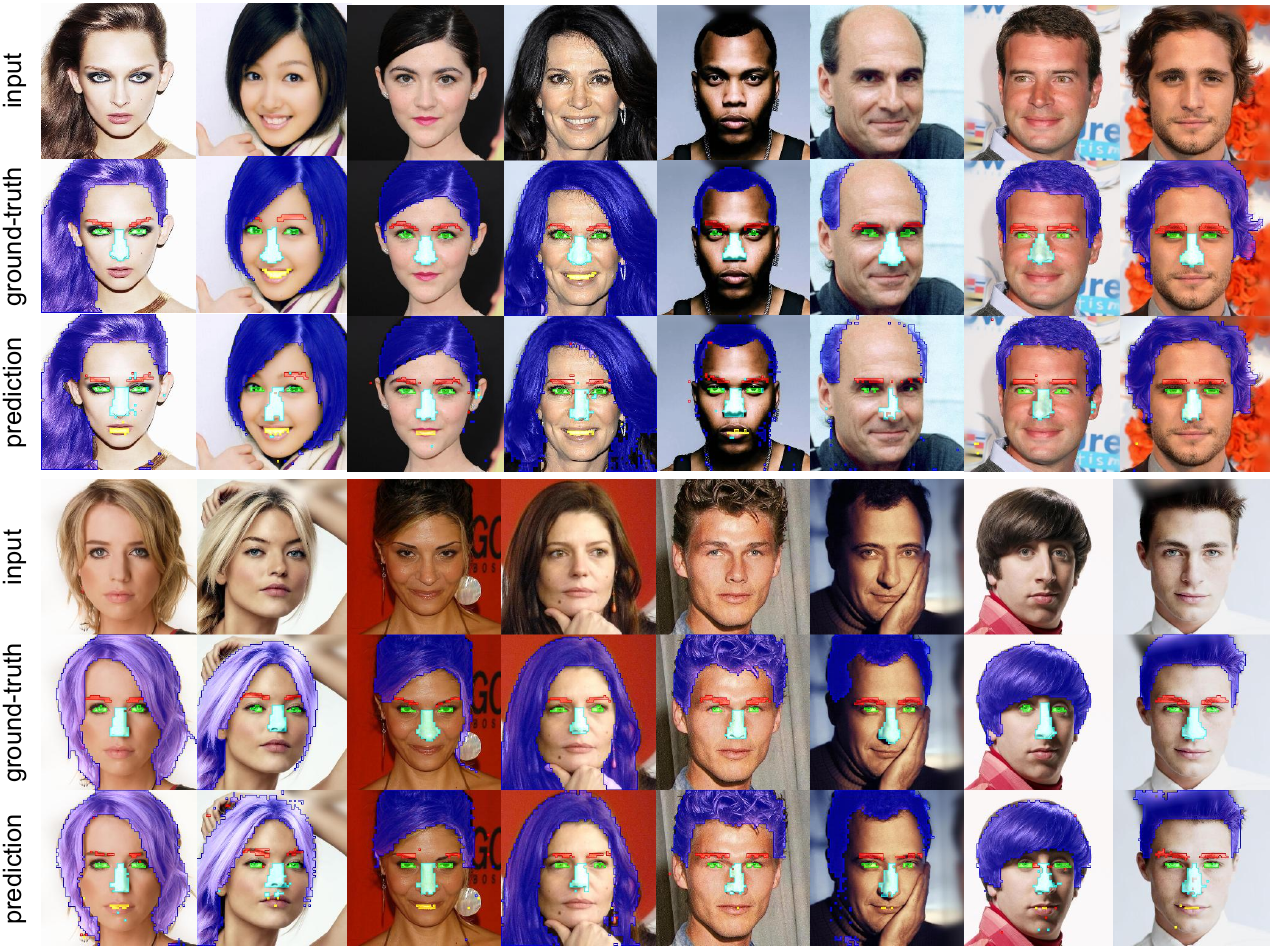}}
\caption{Qualitative results on some randomly sampled test examples from the CelebAMask-HQ dataset.}
\label{fig:app_vis_3}
\end{center}
\end{figure*}

\begin{figure*}
\begin{center}
\centerline{\includegraphics[width=\textwidth]{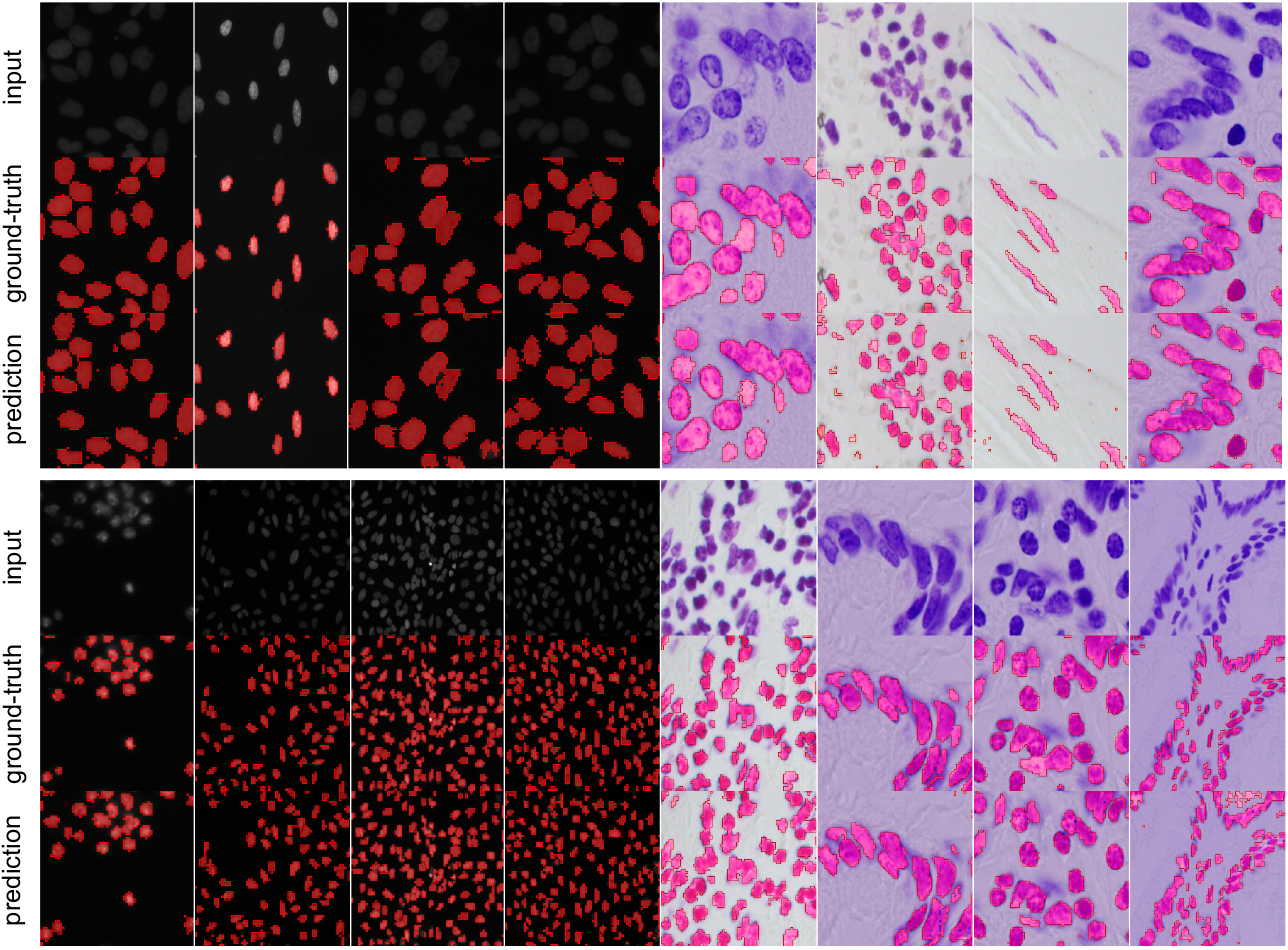}}
\caption{Qualitative results on some randomly sampled test examples from the BBBC038v1 dataset.}
\label{fig:app_vis_4}
\end{center}
\end{figure*}

\begin{figure*}
\begin{center}
\centerline{\includegraphics[width=\textwidth]{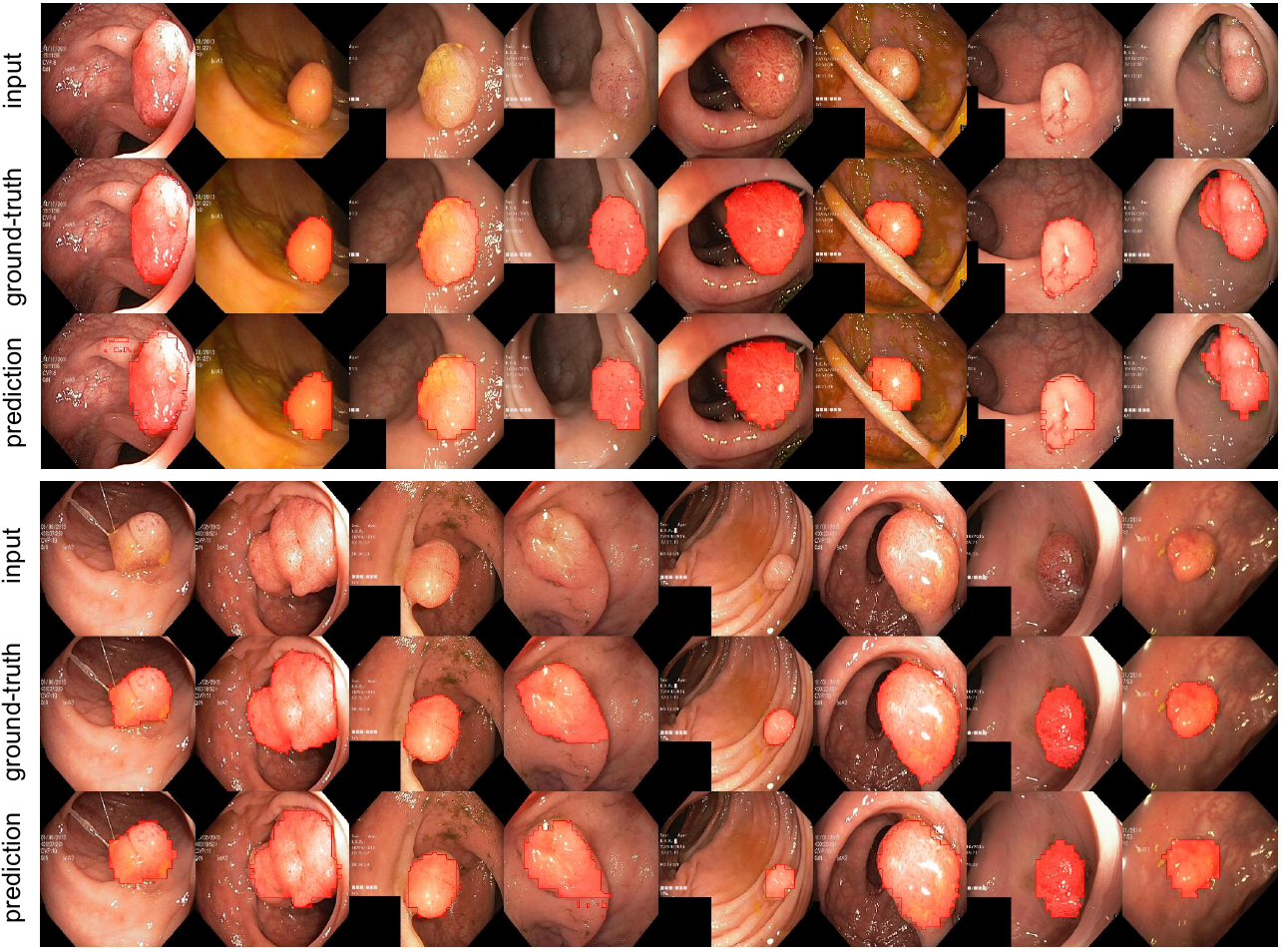}}
\caption{Qualitative results on some randomly sampled test examples from the Kvasir-SEG dataset.}
\label{fig:app_vis_5}
\end{center}
\end{figure*}

\begin{figure*}
\begin{center}
\centerline{\includegraphics[width=\textwidth]{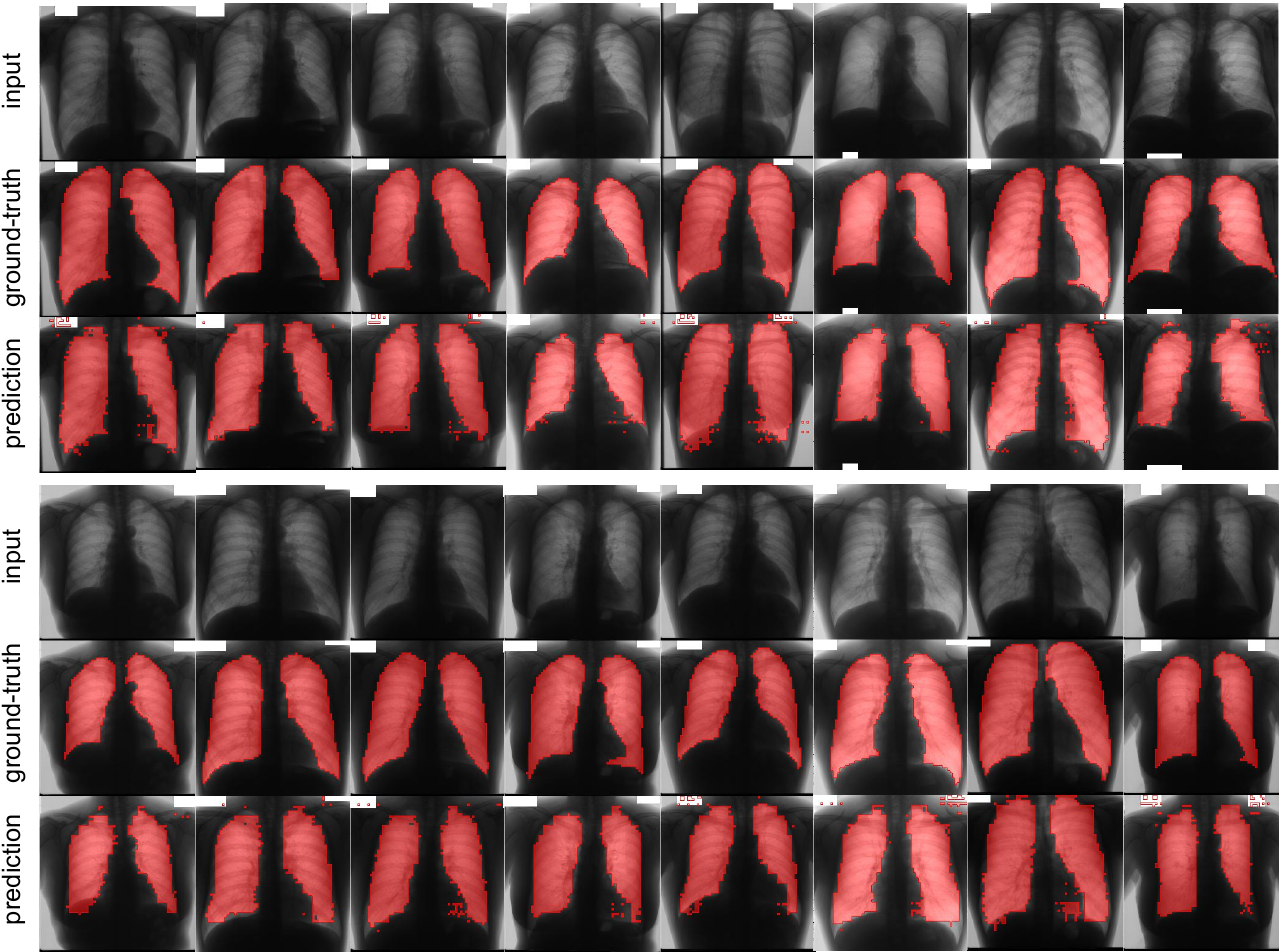}}
\caption{Qualitative results on some randomly sampled test examples from the JSRT dataset.}
\label{fig:app_vis_6}
\end{center}
\end{figure*}

\begin{figure*}
\begin{center}
\centerline{\includegraphics[width=\textwidth]{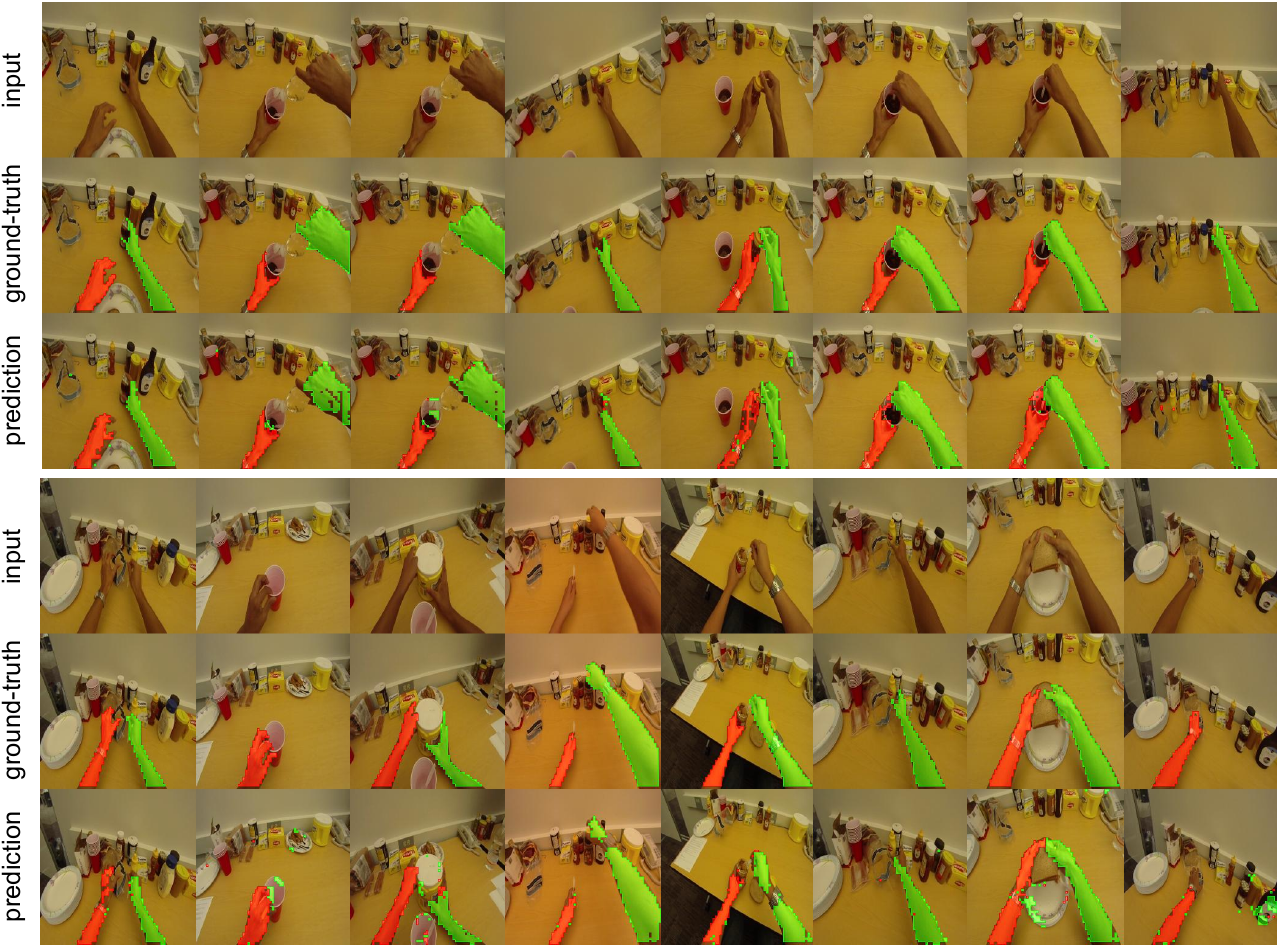}}
\caption{Qualitative results on some randomly sampled test examples from the GTEA dataset.}
\label{fig:app_vis_7}
\end{center}
\end{figure*}

\begin{figure*}
\begin{center}
\centerline{\includegraphics[width=\textwidth]{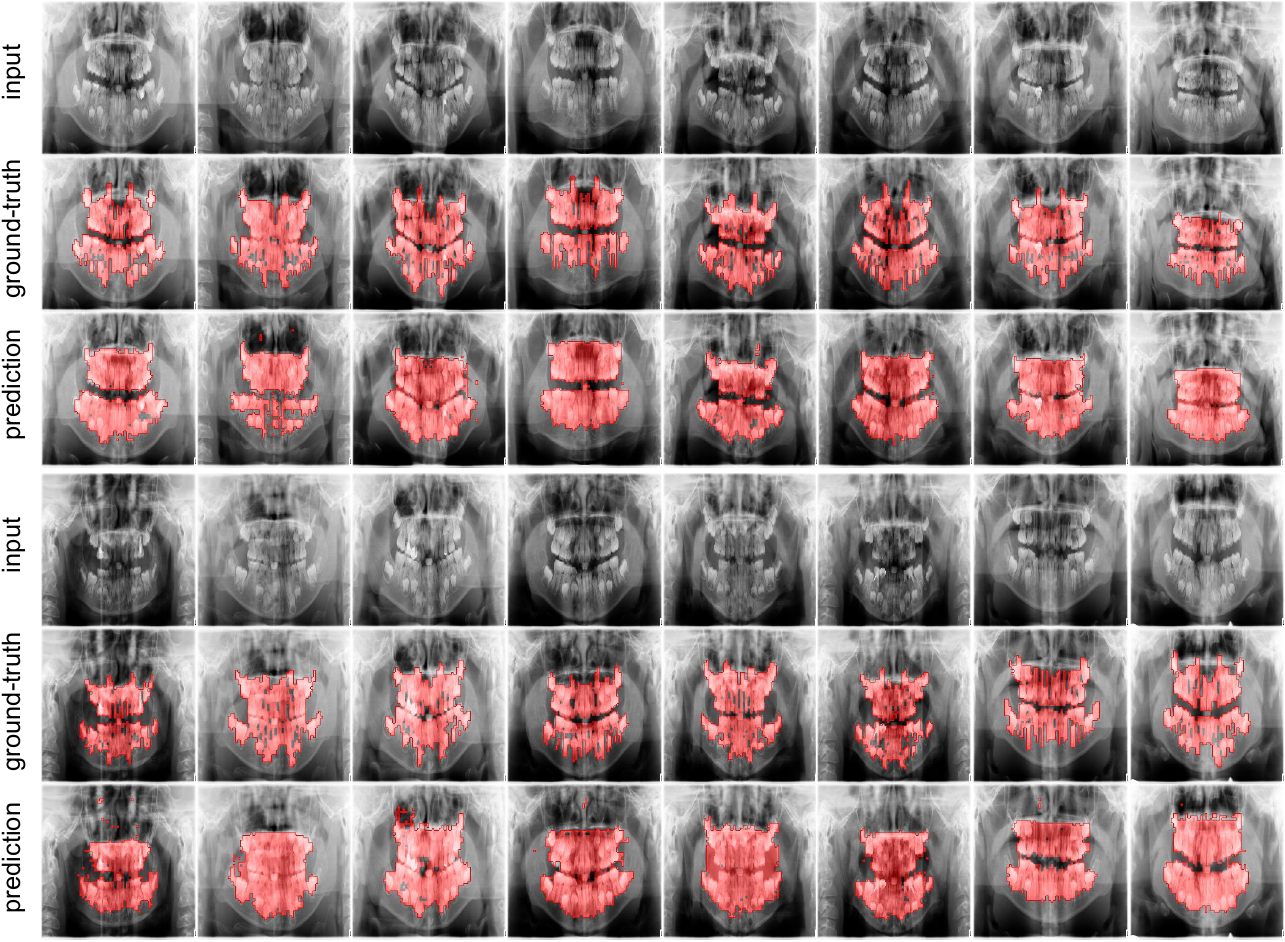}}
\caption{Qualitative results on some randomly sampled test examples from the children’s dental panoramic radiographs dataset.}
\label{fig:app_vis_8}
\end{center}
\end{figure*}

\begin{figure*}
\begin{center}
\centerline{\includegraphics[width=\textwidth]{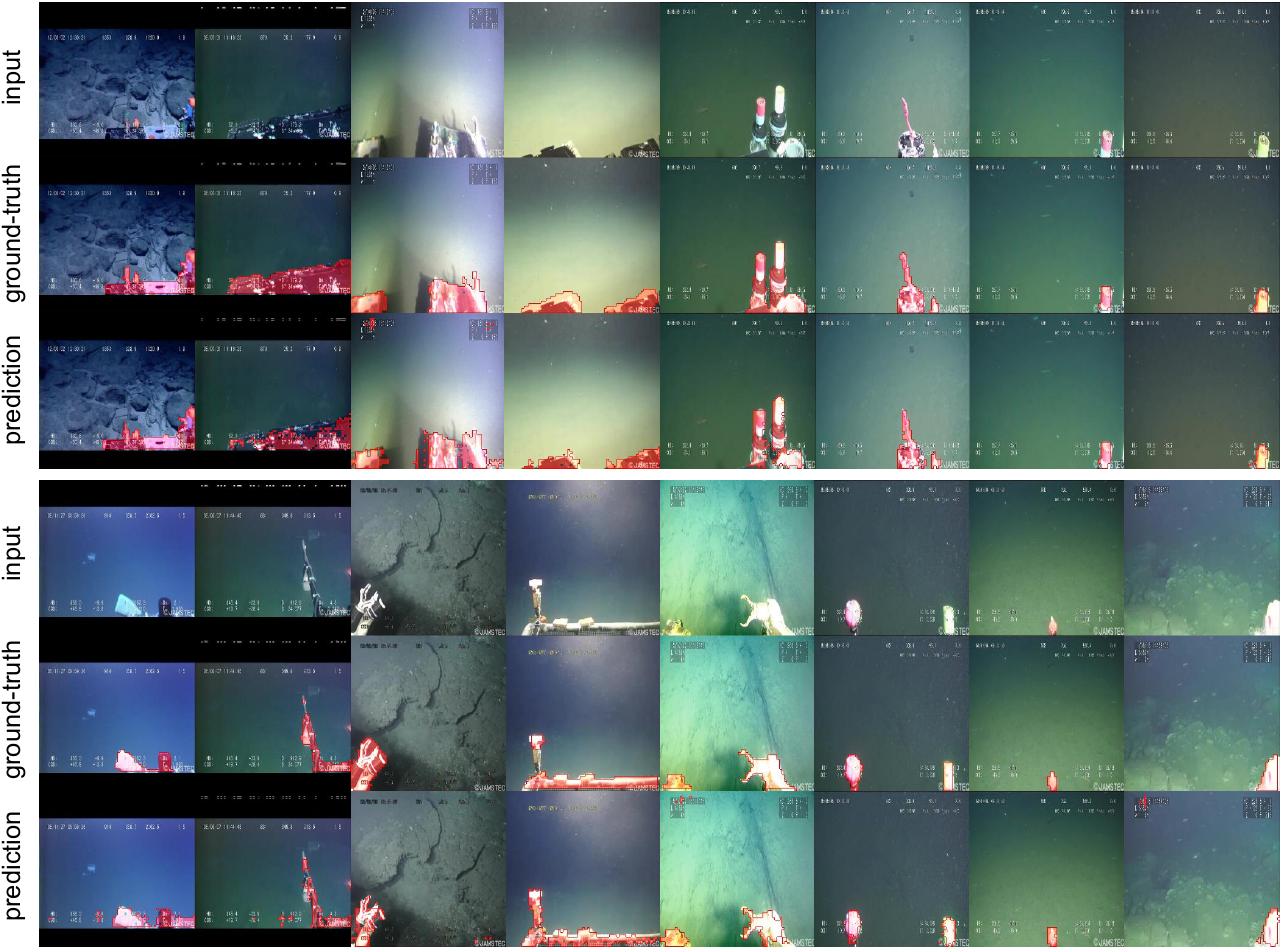}}
\caption{Qualitative results on some randomly sampled test examples from the TrashCa dataset.}
\label{fig:app_vis_9}
\end{center}
\end{figure*}

In Figures \ref{fig:app_vis_1}, \ref{fig:app_vis_2}, \ref{fig:app_vis_3}, \ref{fig:app_vis_4}, \ref{fig:app_vis_5}, \ref{fig:app_vis_6}, \ref{fig:app_vis_7}, \ref{fig:app_vis_8}, and \ref{fig:app_vis_9}, we present the segmentation masks generated by CPC-SAM on different benchmark datasets. In each figure, the first row is the sample, the second row is the ground-truth mask, and the third row is the mask obtained by CPC-SAM. The results demonstrate that across various tasks, the segmentation masks predicted by CPC-SAM consistently exhibit exceptional levels of detail, closely aligning with the ground truth segmentation templates. A noteworthy observation is that CPC-SAM achieves segmentation results in complex OVMS scenarios, such as TrashCan and GTEA, that are almost entirely consistent with the ground truth labels. This indicates that CPC-SAM can achieve robust and accurate OVMS even under few-shot settings. However, in single-entity segmentation tasks, such as TikTok dances, the model may occasionally misclassify individual background pixels as part of the target, an issue not observed in other multi-object segmentation tasks. We hypothesize that this is due to the single-object nature of the task, which leads the model to classify foreground regions near the object boundary as part of the target. Compared with the segmentation results of other baselines, e.g., as shown in \cite{zhangblo}, it can be seen that CPC-SAM's stable mask on this segmentation task still outperforms the baseline method.
The qualitative results highlight CPC-SAM's effectiveness in capturing complex features while maintaining a clearer distinction between foreground and background, emphasizing its robust performance in OVMS tasks across diverse domains.

\end{document}